\documentclass{article}

\PassOptionsToPackage{numbers,sort&compress}{natbib}

\usepackage[preprint]{static/neurips_2026}

\usepackage[utf8]{inputenc} 
\usepackage[T1]{fontenc}    
\usepackage{hyperref}       
\usepackage{url}            
\usepackage{booktabs}       
\usepackage{nicefrac}       
\usepackage{microtype}      

\usepackage{multirow}

\usepackage[textsize=tiny]{todonotes}

\usepackage[german, main=english]{babel}
\usepackage[autostyle]{csquotes}

\usepackage{amsmath}
\usepackage{amssymb} \usepackage{mathtools} \usepackage{bm} \usepackage{dsfont} \usepackage{amsthm} 
\usepackage[acronym]{glossaries-extra}   \GlsXtrEnableEntryUnitCounting      {abbreviation}    {1}    {document}  
\glsdisablehyper

\usepackage[capitalize,noabbrev,english]{cleveref}

\usepackage{xcolor}
\usepackage{enumitem}   \usepackage{xspace}     
\usepackage{siunitx}

\graphicspath{ {./figures/} }
\usepackage[
     format=plain,           labelformat=simple,      labelsep=period,      textformat=simple,           labelfont=footnotesize,sl,      textfont=footnotesize,      ]{caption}
\usepackage{graphicx}
\usepackage{subcaption}
\usepackage{tabularx}   \usepackage{multirow}   \usepackage{makecell}   \usepackage{booktabs}   

\usepackage{pgfplots}
\pgfplotsset{width=10cm,compat=1.8}
\usepgfplotslibrary{groupplots, statistics, fillbetween}
\usetikzlibrary{calc, spy, decorations.markings}

\usepgfplotslibrary{external}

\usepackage{todonotes}

\usepackage{physics}
\usepackage{enumitem}
\usepackage{amsmath,amssymb,amsfonts,pifont}
\usepackage{textcomp}
\usepackage{footnote}
\usepackage{array}    
\usepackage{tikz}
\usepackage{wrapfig} 

\makesavenoteenv{tabular}
\makesavenoteenv{table}
\usepackage{booktabs,bigstrut,rotating}

\usepackage[table]{xcolor}
\usepackage[nolist, nohyperlinks, printonlyused, withpage, smaller]{acronym}
\usepackage{graphicx,import}    
\usepackage{subcaption}  
\usepackage{algorithm}
\usepackage{algorithmic}
\usepackage{enumitem}
\usepackage{tablefootnote}
\usepackage{caption}
\usepackage{subcaption}
\usepackage[most]{tcolorbox}

\newtcolorbox{graybox}{
	colback=gray!20,    
	colframe=gray!20,   
	boxrule=0pt,      
	arc=1pt,            
	left=1pt, right=1pt, top=1pt, bottom=1pt 
}

\newcommand{\conclusion}[1]{\begin{graybox}
		#1
\end{graybox}}

\usepackage[toc]{appendix}
\usepackage{minitoc}  

 \newtheorem{ass}{Assumption}

\newcommand{\cmark}{\textcolor[RGB]{0,0,0}{\scalebox{1.2}{\ding{51}}}}%
\newcommand{\cmarkg}{\textcolor[RGB]{0,120,0}{\scalebox{1.2}{\ding{51}}}}%

\newcommand{\cred}[1]{\textcolor[rgb]{0.85,0,0}{#1}}%
\newcommand{\cgreen}[1]{\textcolor[rgb]{0,0.55,0}{#1}}%
\newcommand{\cblue}[1]{\textcolor[rgb]{0,0,0.65}{#1}}%

\newcommand{\multiline}[1]{\shortstack{#1}}

\newsavebox{\leftcolboxa}
\newsavebox{\leftcolboxb}

\doparttoc 
\faketableofcontents 

\title{Structure-Preserving Gaussian Processes \\ Via Discrete Euler-Lagrange Equations}

\author{%
	Jan-Hendrik~Ewering\\
	Leibniz Universit{\"a}t Hannover\\
	\texttt{\href{mailto:ewering@imes.uni-hannover.de}{ewering@imes.uni-hannover.de}}\\
	\And
	Kathrin~Fla{\ss}kamp\\
	Saarland University\\
	\texttt{\href{mailto:kathrin.flasskamp@uni-saarland.de}{kathrin.flasskamp@uni-saarland.de}}\\
	\And
	Niklas~{Wahlstr{\" o}m}\\
	Uppsala University\\
	\texttt{\href{mailto:niklas.wahlstrom@it.uu.se}{niklas.wahlstrom@it.uu.se}}\\
	\And
	Thomas\,B.~Sch{\"o}n\\
	Uppsala University\\
	\texttt{\href{mailto:thomas.schon@uu.se}{thomas.schon@uu.se}}\\
	\And
	Thomas~Seel\\
	Leibniz Universit{\"a}t Hannover\\
	\texttt{\href{mailto:seel@imes.uni-hannover.de}{seel@imes.uni-hannover.de}}\\
}

\begin{document}
	

\begin{acronym}
	\acro{rnn}[RNN]{Recurrent Neural Network}
	\acroplural{rnn}[RNN]{Recurrent Neural Networks}
	\acro{narx}[NARX-NN]{Nonlinear Autoregressive Exogenous Neural Network}
	\acroplural{narx}[NARX-NN]{Nonlinear Autoregressive Exogenous Neural Networks}
	\acro{gru}[GRU]{{Gated Recurrent Unit}}
	\acroplural{gru}[GRU]{{Gated Recurrent Units}}
	\acro{lstm}[LSTM]{Long Short-Term Memory}
	\acro{ann}[NN]{Neural Network}	
	\acroplural{ann}[NN]{Neural Networks}
	\acro{ffnn}[FFNN]{Feedforward Neural Network}
	\acroplural{ffnn}[FFNN]{Feedforward Neural Networks}
	\acro{pinn}[PINN]{Physics-Informed Neural Network}
	\acroplural{pinn}[PINN]{Physics-Informed Neural Networks}
	\acro{gp}[GP]{Gaussian Process}
    \acro{lgp}[LGP]{Lagrangian Gaussian Process}
    \acro{hgp}[HGP]{Hamiltonian Gaussian Process}
	\acroplural{gp}[GPs]{Gaussian Processes}
    \acroplural{lgp}[LGPs]{Lagrangian Gaussian Processes}
	\acro{knn}[$K$NN]{$K$-Nearest-Neighbors}
	\acro{ilc}[ILC]{Iterative Learning Control}
	\acro{ili}[ILI]{Iterative Learning Identification}
	\acro{rls}[RLS]{Recursive Least Squares}
	\acro{rc}[RC]{Repetitive Control}
	\acro{rl}[RL]{{Reinforcement Learning}}
	\acro{daoc}[DAOC]{{Direct Adaptive Optimal Control}}
	\acro{ml}[ML]{maschinelles Lernen}
	\acro{lwpr}[LWPR]{{Locally Weighted Projection Regression}}
	\acro{svm}[SVM]{{Support Vector Machine}}
	\acro{mcmc}[MCMC]{{Markov Chain Monte Carlo}}
	\acro{ad}[AD]{{Automatic Differentiation}}
	\acro{gmm}[GMM]{{Gaussian Mixture Models}}
	\acro{rkhs}[RKHS]{{Reproducing Kernel Hilbert Spaces}}
	\acro{rbf}[RBF]{{Radial Basis Function}}
	\acro{rbfnn}[RBF-NN]{{Radial Basis Function Neural Network}}
	\acroplural{rbfnn}[RBF-NN]{{Radial Basis Function Neural Networks}}
	\acro{node}[{Neural} ODE]{{Neural Ordinary Differential Equation}}
	\acroplural{node}[{Neural} ODEs]{{Neural Ordinary Differential Equations}}
	\acro{pdf}[PDF]{Probability Density Function}
	\acro{pca}[PCA]{Principal Component Analysis}
	\acro{lnn}[LNN]{Lagrangian Neural Network}
	\acroplural{lnn}[LNNs]{Lagrangian Neural Networks}
	\acro{hnn}[HNN]{Hamiltonian Neural Network}
	\acroplural{hnn}[HNNs]{Hamiltonian Neural Networks}

	\acro{kf}[KF]{Kalman Filter}	
	\acro{ekf}[EKF]{Extended Kalman Filter}
	\acro{nekf}[NEKF]{Neural Extended Kalman Filter}
	\acro{ukf}[UKF]{Unscented Kalman Filter}
	\acro{pf}[PF]{Particle Filter}
	\acro{mhe}[MHE]{{Moving Horizon Estimation}}
	\acro{rpe}[RPE]{{Recursive Predictive Error}}
	\acro{slam}[SLAM]{{Simultaneous Location and Mapping}}
	
	\acro{mftm}[MFTM]{Magic Formula Tire Model}
	\acro{mbs}[MBS]{Multi-Body Simulation}
	\acro{lti}[LTI]{Linear Time-Invariant}
	\acro{cog}[COG]{Center Of Gravity}
	\acro{ltv}[LTV]{Linear Time-Variant}
	\acro{siso}[SISO]{Single Input Single Output}
	\acro{mimo}[MIMO]{Multiple Input Multiple Output}
	\acro{psd}[PSD]{Power Spectral Density}
	\acroplural{psd}[PSD]{Power Spectral Densities}
	\acro{cf}[CF]{Coordinate Frame}
	\acroplural{cf}[CF]{Coordinate Frames}
	\acro{phs}[PHS]{Port-Hamiltonian System}

	\acro{pso}[PSO]{Particle Swarm Optimization}
	\acro{sqp}[SQP]{Sequentielle Quadratische Programmierung}
	\acro{svd}[SVD]{Singular Value Decomposition}
	\acro{ode}[ODE]{Ordinary Differential Equation}
	\acroplural{ode}[ODEs]{Ordinary Differential Equations}
	\acro{pde}[PDE]{Partial Differential Equation}
	
	\acro{nmse}[NMSE]{Normalized Mean Squared Error}
	\acro{mse}[MSE]{Mean Squared Error}
	\acro{rmse}[RMSE]{Root Mean Squared Error}
	\acro{wrmse}[wRMSE]{weighted \ac{rmse}}
	
	\acro{mpc}[MPC]{{Model Predictive Control}}
	\acro{nmpc}[NMPC]{Nonlinear Model Predictive Control}
	\acro{lmpc}[LMPC]{Learning Model Predictive Control}
	\acro{ltvmpc}[LTV-MPC]{\acl{ltv} Model Predictive Control}
	\acro{ndi}[NDI]{Nonlinear Dynamic Inversion}
	\acro{ac}[AC]{Adhesion Control}
	\acro{esc}[ESC]{Electronic Stability Control}
	\acro{ass}[ASS]{Active Suspension System}
	\acro{trc}[TRC]{Traction Control}
	\acro{abs}[ABS]{Anti-Lock Brake System}
	\acro{gcc}[GCC]{Global Chassis Control}
	\acro{ebs}[EBS]{Electronic Braking System}
	\acro{adas}[ADAS]{Advanced Driver Assistance Systems}
	\acro{cm}[CM]{{Condition Monitoring}}
	\acro{hil}[HiL]{{Hardware-in-the-Loop}}
	\acro{siso}[SISO]{Single Input Single Output}
	\acro{mimo}[MIMO]{Multiple Input Multiple Output}

	\acro{irw}[IRW]{Independently Rotating Wheels}
	\acro{dirw}[DIRW]{Driven \acl{irw}}
	\acro{imes}[imes]{{Institute of Mechatronic Systems}}
	\acro{db}[DB]{Deutsche Bahn}
	\acro{ice}[ICE]{Intercity-Express}
	
	\acro{fmi}[FMI]{{Functional Mock-up Interface}}
	\acro{fmu}[FMU]{{Functional Mock-up Unit}}
	\acro{doi}[DOI]{{Digital Object Identifier}}

	\acro{ra}[RA]{Research Area}
	\acroplural{ra}[RA]{Research Areas}
	\acro{wp}[WP]{Work Package}
	\acroplural{wp}[WP]{Work Packages}
	
	\acro{fb}[FB]{Forschungsbereich}
	\acroplural{fb}[FB]{Forschungsbereiche}
	\acro{ap}[AP]{Arbeitspaket}
	\acroplural{ap}[AP]{Arbeitspakete}
	\acro{abb}[Abb.]{Abbildung}
	\acro{luis}[LUIS]{Leibniz Universit"at IT Services}
	
	\acro{res}[RES]{Renewable Energy Sources}
	\acro{pkw}[PKW]{Personenkraftwagen}
	\acro{twipr}[TWIPR]{{Three-Wheeled Inverted Pendulum Robot}}
	\acro{tlr}[TLR]{Two-Link Robot}
	
	\acro{pmcmc}[PMCMC]{Particle Markov Chain Monte Carlo}
	\acro{mcmc}[MCMC]{Markov Chain Monte Carlo}
	\acro{rbpf}[RBPF]{Rao-Blackwellized Particle Filter}
	\acro{pf}[PF]{Particle Filter}
	\acro{ps}[PS]{Particle Smoother}
	\acro{smc}[SMC]{Sequential Monte Carlo}
	\acro{csmc}[cSMC]{conditional SMC}
	\acro{mh}[MH]{Metropolis Hastings}
	\acro{em}[EM]{Expectation Maximization}
	\acro{slam}[SLAM]{Simultaneous Location and Mapping}
	\acro{dof}[DOF]{Degree of Freedom}
	\acroplural{dof}[DOF]{Degrees of Freedom}
	\acro{pg}[PG]{Particle Gibbs}
	\acro{pgas}[PGAS]{Particle Gibbs with Ancestor Sampling}
	\acro{mpgas}[mPGAS]{marginalized Particle Gibbs with Ancestor Sampling}
	\acro{hmm}[HMM]{Hidden Markov Model}
	
	\acro{emps}[EMPS]{Electro-Mechanical Positioning System}
	
	\acro{ilc}[ILC]{Iterative Learning Control}
	\acro{ddilc}[DD-ILC]{Data-Driven Iterative Learning Control}
	\acro{dilc}[DILC]{Dual Iterative Learning Control}
	\acro{iml}[IML]{Iterative Model Learning}
	\acro{noilc}[NO-ILC]{Norm-Optimal Iterative Learning Control}
	\acro{gilc}[G-ILC]{Gradient Iterative Learning Control}

	\acro{bilbo}[BILBO]{Balancing Intelligent Learning roBOt}
    
	\acro{llm}[LLM]{Large Language Model}
    \acroplural{llm}[LLMs]{Large Language Models}
	\acro{ard}[ARD]{Automatic Relevance Determination}

\end{acronym}

\newcommand{\qvec}[1]{\mathbf{#1}}
\newcommand{\qmat}[1]{\mathbf{#1}}
\newcommand{\idx}[1]{_{\mathrm{#1}}}

\newcommand{\qRealNumbers}{\mathbb{R}}
\newcommand{\qPositiveRealNumbers}{\mathbb{R}_{\geq 0}}
\newcommand{\qNaturalNumbersZero}{\mathbb{N}_{\geq 0}}
\newcommand{\qNaturalNumbersPos}{\mathbb{N}_{>0}}
\newcommand{\qNaturalNumbers}{\mathbb{N}}
\newcommand{\foralljinN}{\forall j \in \qNaturalNumbersZero, \quad}
\newcommand{\foralljgeqN}{\forall j \in \qNaturalNumbersPos, \quad}
\newcommand{\qBLT}{\mathcal{T}^{\mathrm{BLT}}}
\newcommand{\qNormal}{\mathcal{N}}

\newcommand\invChi{\mathop{\mbox{Scale-inv-$\chi^2$}}}

\newcommand{\qa}{\qvec{a}}
\newcommand{\qb}{\qvec{b}}
\newcommand{\qd}{\qvec{d}}
\newcommand{\qe}{\qvec{e}}
\newcommand{\qf}{\qvec{f}}
\newcommand{\qg}{\qvec{g}}
\newcommand{\qh}{\qvec{h}}
\newcommand{\qi}{\qvec{i}}
\newcommand{\qj}{\qvec{j}}
\newcommand{\qk}{\qvec{k}}
\newcommand{\ql}{\qvec{l}}
\newcommand{\qm}{\qvec{m}}
\newcommand{\qn}{\qvec{n}}
\newcommand{\qo}{\qvec{o}}
\newcommand{\qp}{\qvec{p}}
\newcommand{\qr}{\qvec{r}}
\newcommand{\qs}{\qvec{s}}
\newcommand{\qt}{\qvec{t}}
\newcommand{\qu}{\qvec{u}}
\newcommand{\qv}{\qvec{v}}
\newcommand{\qw}{\qvec{w}}
\newcommand{\qx}{\qvec{x}}
\newcommand{\qy}{\qvec{y}}
\newcommand{\qz}{\qvec{z}}
\newcommand{\qem}{\qvec{e}^{\mathrm{m}}}

\newcommand{\qA}{\qvec{A}}
\newcommand{\qB}{\qvec{B}}
\newcommand{\qC}{\qvec{C}}
\newcommand{\qD}{\qvec{D}}
\newcommand{\qE}{\qvec{E}}
\newcommand{\qF}{\qvec{F}}
\newcommand{\qG}{\qvec{G}}
\newcommand{\qH}{\qvec{H}}
\newcommand{\qI}{\qvec{I}}
\newcommand{\qJ}{\qvec{J}}
\newcommand{\qK}{\qvec{K}}
\newcommand{\qL}{\qvec{L}}
\newcommand{\qM}{\qvec{M}}
\newcommand{\qN}{\qvec{N}}
\newcommand{\qO}{\qvec{O}}
\newcommand{\qP}{\qvec{P}}
\newcommand{\qQ}{\qvec{Q}}
\newcommand{\qR}{\qvec{R}}
\newcommand{\qS}{\qvec{S}}
\newcommand{\qT}{\qvec{T}}
\newcommand{\qU}{\qvec{U}}
\newcommand{\qV}{\qvec{V}}
\newcommand{\qW}{\qvec{W}}
\newcommand{\qX}{\qvec{X}}
\newcommand{\qY}{\qvec{Y}}
\newcommand{\qZ}{\qvec{Z}}

\newcommand{\quv}{\bar{\qu}}
\newcommand{\qyv}{\bar{\qy}}
\newcommand{\qxv}{\bar{\qx}}
\newcommand{\qZero}{\qvec{0}}
\newcommand{\qdu}{\boldsymbol{\Delta}\qu}
\newcommand{\qdy}{\boldsymbol{\Delta}\qy}
\newcommand{\qep}{\hat{\qe}}
\newcommand{\qyp}{\hat{\qy}}

\newcommand{\qILCDesign}{\boldsymbol{D}}
\newcommand{\qIMLDesign}{\boldsymbol{\hat{D}}}
\newcommand{\qLIML}{\hat{\qL}}
\newcommand{\qeR}{\qe_\mathrm{R}}


\newcommand{\qff}{\boldsymbol{f}}
\newcommand{\qpf}{\boldsymbol{p}}
\newcommand{\qmf}{\boldsymbol{m}}
\newcommand{\qXf}{\boldsymbol{A}}

\newcommand{\qLift}{\mathcal{L}}
\newcommand{\qLu}{\boldsymbol{\mathcal{L}\idx{u}}}
\newcommand{\qLm}{\boldsymbol{\mathcal{L}\idx{m}}}
\newcommand{\qUb}{\bar{\qU}}
\newcommand{\qyh}{\hat{\qy}}
\newcommand{\qeh}{\hat{\qe}}
\newcommand{\qLh}{\hat{\qL}}
\newcommand{\qDeh}{\hat{\boldsymbol{\mathcal{D}}}}
\newcommand{\qDe}{{\boldsymbol{\mathcal{D}}}}
\newcommand{\qMt}{\tilde{\qM}}
\newcommand{\qPt}{\tilde{\qP}}
\newcommand{\qLt}{\tilde{\qL}}
\newcommand{\qQt}{\tilde{\qQ}}
\newcommand{\qSt}{\tilde{\qS}}
\newcommand{\qWt}{\tilde{\qW}}
\newcommand{\qRt}{\tilde{\qR}}
\newcommand{\qut}{\tilde{\qu}}
\newcommand{\qyt}{\tilde{\qy}}
\newcommand{\qet}{\tilde{\qe}}
\newcommand{\qzt}{\tilde{\qz}}
\newcommand{\qrt}{\tilde{\qr}}
\newcommand{\qXt}{\tilde{\qX}}

\newcommand{\qTb}{\bar{\qT}}
\newcommand{\qVb}{\bar{\qV}}

\newcommand{\qQh}{\hat{\qQ}}
\newcommand{\qSh}{\hat{\qS}}
\newcommand{\qXh}{\hat{\qX}}
\newcommand{\qWh}{\hat{\qW}}

\newcommand{\Lfunc}{\mathscr{L}}       
\newcommand{\Kfunc}{\mathscr{K}}
\newcommand{\KLfunc}{\Kfunc\negthinspace\negthinspace\Lfunc}

\newcommand{\qTwoNorm}[1]{\left\| {#1} \right\|_{2}}
\newcommand{\qInfNorm}[1]{\left\| {#1} \right\|_\infty}
\newcommand{\qOneNorm}[1]{\left\| {#1} \right\|\idx{1}}
\newcommand{\qNorm}[1]{\left\| {#1} \right\|}
\newcommand{\qGivenNorm}{\qNorm{\boldsymbol{\cdot}}}

\newcommand{\qred}[1]{{\color{red}#1}}
\newcommand{\imesorange}{E77B29}
\newcommand{\imesgruen}{C8D317}
\newcommand{\imesblauHundert}{00509B}
\newcommand{\imesblauZwanzig}{CCDCEB}
\newcommand{\imesblauVierzig}{99B9D8}

\newcommand{\eg}{e.\,g.,\,}
\newcommand{\ie}{i.\,e.,\,}
\newcommand*{\R}{\mathbb{R}}

\newcommand*{\MNIW}{\mathcal{MNIW}}
\newcommand*{\IW}{\mathcal{IW}}

\newcommand{\del}{\partial}
\newcommand{\bi}[1]{\boldsymbol{#1}}
\newcommand{\ur}[1]{\mathrm{#1}}
\newcommand{\cali}[1]{\mathcal{#1}}

\newcommand{\ubar}[1]{\underaccent{\bar}{#1}}
\newcommand{\ToDo}[1]{\todo[size=\tiny]{#1}}

\newcommand{\spans}[1]{\spanop\left( #1 \right)}

\newcommand{\ToDos}[1]{\todo[size=\tiny]{#1}}


	\maketitle

	\begin{abstract}
        %
		In this paper, we propose \acp{lgp} for probabilistic and data-efficient learning of dynamics via discrete forced Euler-Lagrange equations. 
		Importantly, the geometric structure of the Lagrange-d'Alembert principle, which governs the motion of dynamical systems, is preserved by construction in the absence of external forces. 
		This allows learning physically consistent models that overcome erroneous drift in the system's energy, thereby providing stable long-term predictions. 
		At the core of our approach lie linear operators for Gaussian process conditioning, constructed from discrete forced Euler-Lagrange equations and variational discretization schemes. 
		Thereby and unlike prior work, the method enables learning dynamics from \textit{discrete position snapshots}, \ie without access to a system's velocities or momenta.
		This is particularly relevant for a large class of practical scenarios where only position measurements are available, for instance, in motion capture or visual servoing applications.  
		We demonstrate the data-efficiency and generalization capabilities of the \acp{lgp} in various synthetic and real-world case studies, including a real-world soft robot with hysteresis.
		The experimental results underscore that the \acp{lgp} learn physically consistent dynamics with uncertainty quantification solely from sparse positional data and enable stable long-term predictions.
        %
	\end{abstract}

	\section{Introduction}
	Exploiting known physical principles when learning dynamics models has become a key strategy for improving data efficiency and generalization \cite{Watson.2025}. 
	In this regard, learning physically consistent models from data can enable the use of well-known, reliable model-based planning or control in complex real-world applications, such as fluid mechanics \cite{Raissi.2020} or soft robotics \cite{Liu.2024,Habich.2026}. 
	
	A powerful approach to enable efficient learning without losing expressivity is to incorporate only a very general system understanding. 
	One example of such \textit{non-restrictive} prior knowledge is the Lagrange-d'Alembert principle, which governs the motion of dynamical systems via a foundational energy law. 
	Seminal work on so-called \acp{lnn} \cite{Lutter.2019b} or \acp{hnn} \cite{Greydanus.2019} leverage this energy principle as algebraic inductive biases to enforce hard physical constraints at minimal computational cost. 
	This approach contrasts with sampling-heavy collocation methods for physics-informed learning \cite{Raissi.2019,Raissi.2020}. 

	Recent work extends \cite{Lutter.2019b,Greydanus.2019} to enable energy-consistent learning in practically relevant scenarios \cite{Liu.2024,Weiss.2026} by considering non-canonical coordinates \cite{Cranmer.2019} or non-conservative systems \cite{Xiao.2024}, \ie systems with energy dissipation and/or control inputs. 
	However, this prior work is limited in three directions: \textit{(i)} structural preservation of the underlying energy conservation law, \textit{(ii)} uncertainty quantification, and \textit{(iii)} learning from position data. 
	
	\textit{First}, providing physically \emph{consistent} long-term predictions---without erroneous drift of the system's energy---needs special attention. 
	Namely, the dynamics learning and prediction scheme needs to preserve the geometric structure of the underlying energy principle \cite{OberBlobaum.2011}. 
	In this regard, most existing work requires specific symplectic integration schemes or ignores structure preservation altogether. 
	
	\textit{Second}, an uncertainty quantification is highly relevant for various applications, such as stochastic control or safe learning \citep{Brunke.2022}. 
	However, the few works that propose learning mechanisms for preserving the geometric structure of the energy principle do not provide {probabilistic} dynamics models \cite{Hansen.2025,Li.2026}.

	\textit{Third}, the vast majority of existing work on energy-consistent learning relies on measurements of the entire system state \cite{Beckers.2022,Dai.2024,Giacomuzzo.2024}. 
	This means that, \eg beyond position snapshots, momentum or velocity measurements are required for training, which is a restrictive assumption in many practical settings.

	\begin{figure}[tb]
		\centering
		{\fontsize{8pt}{8pt}\selectfont
			\resizebox{1\textwidth}{!}{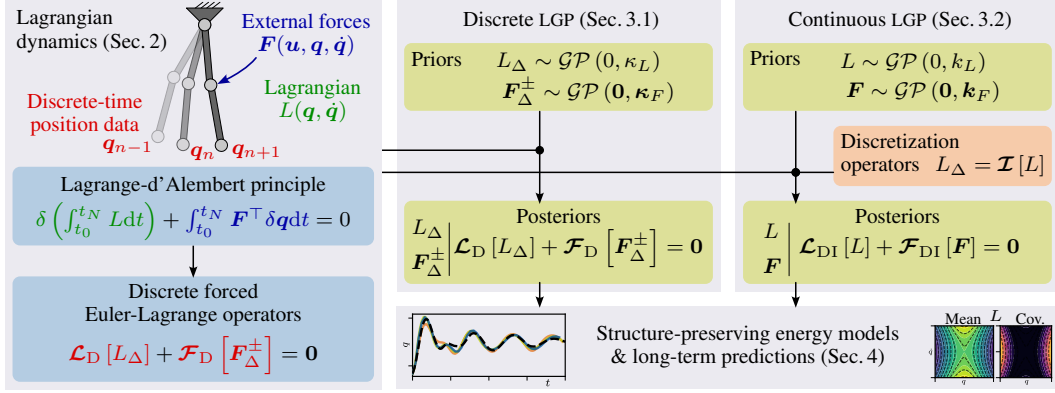}}
		\caption{We propose \aclp{lgp} (\acsp{lgp}) for learning probabilistic, non-conservative dynamics models using only position measurements, without access to velocity or momentum data. Incorporating Lagrange-d'Alembert principle into \acp{gp}, the methods enable physically consistent long-term predictions.}
		\label{fig:method_overview}
		\vspace{-2mm}
	\end{figure}
	
	In this paper, we propose structure-preserving \aclp{lgp} (\acsp{lgp}) for learning probabilistic non-conservative dynamics models. 
    We stress that the \acp{lgp} do not require system-specific prior knowledge but use only the Lagrange-d'Alembert principle. 
	By harnessing discrete forced Euler-Lagrange linear operators, we are able to learn the \acp{gp} only from position data, without requiring restrictive velocity or momentum measurements (see Figure\,\ref{fig:method_overview}). 
	We contextualize our approach to related work in Section\,\ref{sec:rel_works}, and present an overview in Table\,\ref{tab:lit_rev}. 
	Our contributions are:

	\begin{enumerate}[label=\textit{(\alph*)}, leftmargin=*, nosep]
		\item Two structure-preserving \ac{lgp} schemes, a discrete and a continuous version, for probabilistic learning of dynamics models, characterized by the system's Lagrangian and external force functions. 
		\item A validation in various synthetic and real-world systems, including a controlled double pendulum and a controlled pneumatic soft robot. 
	\end{enumerate}
	
	Specifically, both the discrete and continuous \acp{lgp} preserve the symplectic structure of the underlying energy principle, thereby enabling energy conservation and stable long-term predictions. 
	By using a normalization condition, our approach is not restricted to a fixed kernel choice, unlike most prior work. 
	Yet, if additional knowledge about the energy model is available, it can be incorporated into the \ac{gp} kernel to facilitate learning. 
	The continuous \ac{lgp} additionally generalizes to custom prediction time-step widths and allows constructing functions for linear observables of the learned Lagrangian, such as the Hamiltonian or the conjugate momentum.

	\begin{table}[htbp]
		\vspace{-2mm}
		\centering
		\caption{Recent work on learning energy-based dynamics models. A system's energy function (Hamiltonian) is denoted $H$, and its Lagrangian function is $L$.}
		{\fontsize{8pt}{8pt}\selectfont
			\resizebox{1\linewidth}{!}{
				\begin{tabular}{lcccccccccccc}
					\toprule
					&       & \multicolumn{3}{c}{\textbf{Neural networks}} &       & \multicolumn{7}{c}{\textbf{Gaussian processes}} \\
					& \multicolumn{1}{l}{~} & \multicolumn{1}{l}{\cite{Greydanus.2019}} & \multicolumn{1}{l}{\cite{Cranmer.2019}} & \multicolumn{1}{l}{\cite{Hansen.2025}} & \multicolumn{1}{l}{~} & \multicolumn{1}{l}{\cite{Tanaka.2022}} & \multicolumn{1}{l}{\cite{Beckers.2022}} & \multicolumn{1}{l}{\cite{Ewering.2025}} & \multicolumn{1}{l}{\cite{OberBlobaum.2023,Offen.2025}} & \multicolumn{1}{l}{\cite{Evangelisti.2022b}} & \multicolumn{1}{l}{\cite{Dai.2024,Giacomuzzo.2024}} & \multicolumn{1}{l}{Proposed} \\
					Energetic quantity &       & $H$   & $L$   & $L$   &       & $H$   & $H$   &   $H$     & $L$   & $L$   & $L$   & $L$ \\
					\cmidrule{1-1}\cmidrule{3-5}\cmidrule{7-13}        {\textit{(i)} Structure-preserving by construction} &       &       &       & \cmark &       &       &       &       & \cmark &       &       & \cmarkg \\
					{\textit{(ii)} Uncertainty quantification} &       &       &       &       &       & \cmark & \cmark & \cmark & \cmark & \cmark & \cmark & \cmarkg \\
					{\textit{(iii)} Learns only from positions} &       &       &       & \cmark &       &       &       & \cmark & \cmark &       &       & \cmarkg \\
					{Consider dissipation \& inputs} &       &       &       & $\sim$\footnote{This approach considers \textit{either} dissipation \textit{or} external control inputs to the system.\label{fnt:shared}} &       & $\sim$\textsuperscript{\ref{fnt:shared}} & \cmark & \cmark &       & $\sim$\textsuperscript{\ref{fnt:shared}} & \cmark & \cmarkg \\
					\bottomrule
				\end{tabular}%
		}}
		\label{tab:lit_rev}%
	\end{table}%

	\section{Background}\label{sec:background}

	\paragraph{Continuous Lagrange-d'Alembert Principle.}
	The motion of dynamical systems with non-conservative forces, \ie that have driving inputs and/or dissipative elements, can be described by the Lagrange-d'Alembert principle. 
	This formulation states that, for a motion path $\boldsymbol{q} : [t_0, t_N] \rightarrow Q \subset \mathbb{R}^{n_q}$, the variation of the system's action plus the virtual work done by external forces must be zero. 
	Here, $Q$ is the configuration manifold that the generalized coordinates $\boldsymbol{q}$ can attain, and $TQ$ is its tangent bundle, describing the set of reachable coordinates $[\boldsymbol{q}, \dot{\boldsymbol{q}}]$.  
	Formally, the Lagrange-d'Alembert principle reads
	\begin{equation}\label{eq:lagrange_dalembert}
		\cgreen{ \delta \left( \int_{t_0}^{t_N} L(\boldsymbol{q}(t), \dot{\boldsymbol{q}}(t)) \mathrm{d} t \right) } + \cblue{\int_{t_0}^{t_N} \boldsymbol{F}(\boldsymbol{u}(t), \boldsymbol{q}(t), \dot{\boldsymbol{q}}(t))^{\top} \delta \boldsymbol{q}(t) \mathrm{d} t } = 0  \, ,
	\end{equation}
	where $\delta \boldsymbol{q}$ represents variations that are zero at the path endpoints, \ie $\delta \boldsymbol{q}(t_0) = \delta \boldsymbol{q}(t_N) = \boldsymbol{0}$ \cite{OberBlobaum.2011}. 
	Potential control inputs that drive the system are denoted $\boldsymbol{u} : [t_0, t_N] \rightarrow U \subset \mathbb{R}^{n_u}$. 
	In \eqref{eq:lagrange_dalembert}, the Lagrangian is $L: TQ \rightarrow \mathbb{R}$, and the Lagrangian control force is $\boldsymbol{F}: U \times TQ  \rightarrow T^* Q$, where the cotangent bundle $T^*Q$ represents the space of generalized forces \cite{OberBlobaum.2011}.
	If $\boldsymbol{F}(\cdot) = \boldsymbol{0}$, a system is said to be \textit{conservative}. 
	An equivalent formulation to \eqref{eq:lagrange_dalembert} is provided by the continuous forced Euler-Lagrange \cite{OberBlobaum.2011} equation
	\begin{equation}\label{eq:contin_euler_lagrange}
		\frac{\partial L}{\partial \boldsymbol{q}} (\boldsymbol{q}, \dot{\boldsymbol{q}}) -\frac{\mathrm{d}}{\mathrm{d} t} \left( \frac{\partial L}{\partial \dot{\boldsymbol{q}}} ({{\boldsymbol{q}}}, \dot{\boldsymbol{q}}) \right) + \boldsymbol{F} (\boldsymbol{u}, \boldsymbol{q}, \dot{\boldsymbol{q}}) = \boldsymbol{0} \, ,
	\end{equation}
	which is linear in the Lagrangian $L$ and the external force $\boldsymbol{F}$. 
	This linearity comes in handy for later \ac{gp} conditioning. 
	Alternatively, the system dynamics can be described using forced Hamiltonian mechanics \cite{OberBlobaum.2011}. 
	Specifically, the Hamiltonian $H$, describing a system's total energy, relates to the regular Lagrangian of a given system via a linear operator, the Legendre transform
	\begin{equation}\label{eq:hamiltonian_legendre}
		H (\boldsymbol{q}, \dot{\boldsymbol{q}}) = \dot{\boldsymbol{q}}^{\top} \frac{\partial L}{\partial \dot{\boldsymbol{q}}}(\boldsymbol{q}, \dot{\boldsymbol{q}}) - L (\boldsymbol{q}, \dot{\boldsymbol{q}}) \, .
	\end{equation}
	While various related works build on continuous Euler-Lagrange \cite{Xiao.2024,Giacomuzzo.2024,Evangelisti.2022b,Dai.2024,Cranmer.2019, Lutter.2019b} or Hamiltonian mechanics \cite{Beckers.2022,Tanaka.2022}, all of these formulations rely on velocity or momentum measurements, which are rarely available in practice. 
	Therefore, in the following, we consider the discrete Lagrange-d'Alembert principle, which depends solely on control input and position data $\boldsymbol{q}$. 
	
	\paragraph{Discrete Lagrange-d’Alembert Principle.}
	For applying the Lagrange-d’Alembert principle to discrete data, we must approximate the continuous integrals. 
	To this end, the time interval $[t_0, t_N]$ is divided into $N$ steps of size $h$. 
	The action integral over a small segment $[t_n, t_{n+1}]$, \ie the first part of \eqref{eq:lagrange_dalembert}, is approximated by the discrete Lagrangian $L_{\Delta} : Q \times Q \rightarrow \mathbb{R}$ with
	\begin{equation}\label{eq:discrete_lagrangian}
		\cgreen{ \int_{t_n}^{t_{n+1}} L(\boldsymbol{q}(t), \dot{\boldsymbol{q}}(t)) \mathrm{d} t \approx L_{\Delta}(\boldsymbol{q}_n, \boldsymbol{q}_{n+1}) } \, \qquad \text{and} \qquad  \boldsymbol{q}_n := \boldsymbol{q}(t_n) \, .
	\end{equation}
	Analogously, the virtual work---that is the second part in \eqref{eq:lagrange_dalembert}---is composed of the \textit{left} and \textit{right} discrete force $\boldsymbol{F}_{\Delta}^- : U \times Q \times Q \rightarrow T^* Q$ and $ \boldsymbol{F}_{\Delta}^+ : U \times Q \times Q \rightarrow T^* Q$, respectively, \ie
	\begin{equation}\label{eq:discrete_force}
		\cblue{ \int_{t_n}^{t_{n+1}} \boldsymbol{F}^{\top} \delta \boldsymbol{q}(t) \mathrm{d} t = \underbrace{\left(\int_{t_n}^{t_{n+1}} \boldsymbol{F}^{\top} \frac{\partial \boldsymbol{q}(t)}{\partial \boldsymbol{q}_n} \mathrm{d} t\right)}_{\approx \boldsymbol{F}_{\Delta}^-(\boldsymbol{u}_n,\boldsymbol{q}_n, \boldsymbol{q}_{n+1})^{\top}} \cdot \delta \boldsymbol{q}_n + \underbrace{\left(\int_{t_n}^{t_{n+1}} \boldsymbol{F}^{\top} \frac{\partial \boldsymbol{q}(t)}{\partial \boldsymbol{q}_{n+1}} \mathrm{d} t\right)}_{\approx \boldsymbol{F}_{\Delta}^+(\boldsymbol{u}_n,  \boldsymbol{q}_n, \boldsymbol{q}_{n+1})^{\top}} \cdot \delta \boldsymbol{q}_{n+1} } \, ,
	\end{equation}
	where we skipped some arguments to improve readability. 
	The discrete force $\boldsymbol{F}_{\Delta}^-$ and $\boldsymbol{F}_{\Delta}^+$can be interpreted as the contributions of the continuous force to the virtual work associated with the variation of the left and right endpoints of the path segment \cite{OberBlobaum.2011}. 
	
	Now considering the entire interval $[t_0,t_N]$ again, we can replace the integrals in \eqref{eq:lagrange_dalembert} with the approximations \eqref{eq:discrete_lagrangian} and \eqref{eq:discrete_force} and arrive, after some steps, at the discrete Lagrange-d'Alembert principle
	\begin{equation}\label{eq:discrete_lagrange_dalembert}
		\cgreen{\delta \sum_{n=0}^{N-1} L_{\Delta}(\boldsymbol{q}_n, \boldsymbol{q}_{n+1})} + \cblue{\sum_{n=0}^{N-1} \left[ \boldsymbol{F}_{\Delta}^-(\boldsymbol{u}_n,  \boldsymbol{q}_n, \boldsymbol{q}_{n+1})^{\top} \delta \boldsymbol{q}_n + \boldsymbol{F}_{\Delta}^+(\boldsymbol{u}_n, \boldsymbol{q}_n, \boldsymbol{q}_{n+1})^{\top} \delta \boldsymbol{q}_{n+1} \right] } = 0  \, ,
	\end{equation}
	for all variations $\{\delta \boldsymbol{q}_n\}_{n=0}^N$ vanishing at the endpoints, \ie $\delta \boldsymbol{q}_0 = \delta \boldsymbol{q}_N = \boldsymbol{0}$   \cite{OberBlobaum.2011}. 
	Equivalently, we can write \eqref{eq:discrete_lagrange_dalembert} as the \textit{discrete} forced Euler-Lagrange equations, $\forall n \in [1,\hdots,N-1]$,
	\begin{equation}\label{eq:DEL}
		\nabla_2 L_{\Delta}(\boldsymbol{q}_{n-1}, \boldsymbol{q}_n) + \nabla_1 L_{\Delta}(\boldsymbol{q}_n, \boldsymbol{q}_{n+1}) + \boldsymbol{F}_{\Delta}^+(\boldsymbol{u}_{n-1}, \boldsymbol{q}_{n-1}, \boldsymbol{q}_n) + \boldsymbol{F}_{\Delta}^-(\boldsymbol{u}_{n},  \boldsymbol{q}_n, \boldsymbol{q}_{n+1}) = \boldsymbol{0} \, ,
	\end{equation}
	where $\nabla_i$ denotes the partial derivative w.r.t. the $i$-th argument of a function. 
	
	\section{Non-Conservative Lagrangian Gaussian Processes}\label{sec:methods}
	We introduce two structure-preserving \acp{lgp} that exploit the discrete forced Euler-Lagrange equations for dynamics learning without requiring system-specific prior knowledge. 
	Section\,\ref{sec:dt_lgp} introduces a scheme that learns discrete Lagrangians $L_{\Delta}$ and external forces $\boldsymbol{F}^{\pm}_{\Delta}$. 
	In Section\,\ref{sec:ct_lgp}, we propose a second \ac{lgp} for learning continuous Lagrangians and forces, based on variational discretization.

	\subsection{Method 1: Learning Discrete Lagrangians and Forces}\label{sec:dt_lgp}
	
	\paragraph{Priors.} 
	We model the discrete Lagrangian and external forces with zero-mean \ac{gp} priors
		\begin{align}\label{eq:dlgp_priors}
			L_{\Delta} &\sim \mathcal{G} \mathcal{P}\left(0, {\kappa}_L\left(\boldsymbol{r}, \boldsymbol{r}^{\prime}\right)\right)\, , \qquad \qquad \boldsymbol{F}_{\Delta}^{\pm} \sim \mathcal{G} \mathcal{P}\left(\boldsymbol{0}, \boldsymbol{\kappa}_{F} \left(\boldsymbol{s}, \boldsymbol{s}^{\prime}\right) \right)\, ,
		\end{align}
	where the variables $\boldsymbol{r}_n = \{ \boldsymbol{q}_{n}, {\boldsymbol{q}}_{n+1} \} $ and $\boldsymbol{s}_n = \{ \boldsymbol{u}_{n},  \boldsymbol{q}_{n}, {\boldsymbol{q}}_{n+1} \}$ summarize the arguments in \eqref{eq:discrete_lagrangian} and \eqref{eq:discrete_force} at time step $n$, respectively.
	In particular, we model the left and right external forces $\boldsymbol{F}_{\Delta}^{+}$ and $\boldsymbol{F}_{\Delta}^{-}$ with the same multi-output \ac{gp} as $\boldsymbol{F}_{\Delta}^{\pm}$. 
	Specifically, we choose $\boldsymbol{\kappa}_{F} \left(\boldsymbol{s}, \boldsymbol{s}^{\prime}\right) = \mathbf{I} \otimes \bar{\kappa}_{F}(\boldsymbol{s},\boldsymbol{s}^\prime)$, where $\mathbf{I}$ is the identity matrix, $\otimes$ is the tensor product, and $\bar{\kappa}_{F}(\boldsymbol{s},\boldsymbol{s}^\prime)$ as well as ${\kappa}_L(\boldsymbol{r},\boldsymbol{r}^\prime)$ are valid differentiable scalar kernels, \eg squared exponential kernels. 
	Extensions using coregionalization \cite{Bonilla.2007} can be applied straightforwardly.

	\paragraph{Linear operators.} 
	To condition the \acp{gp} on discrete forced Euler-Lagrange equations, we exploit the fact that \acp{gp} are closed under linear operations \cite{Pfortner.2022}. 
	We define the linear operators $\boldsymbol{\mathcal{L}}_{\mathrm{D}}$, acting on the discrete Lagrangian $L_{\Delta}$, and $\boldsymbol{\mathcal{F}}_{\mathrm{D}}$, acting on the force components $\boldsymbol{F}_{\Delta}^{\pm}$, as
	\begin{subequations}\label{eq:dlgp_lin_operators}
		\begin{align}
			\boldsymbol{\mathcal{L}}_{\mathrm{D}}\left[L_{\Delta}\right](\boldsymbol{q}_a, \boldsymbol{q}_b, \boldsymbol{q}_c) 
			&\triangleq 
			\left.\nabla_{2} L_{\Delta} \right|_{\boldsymbol{q}_a, \boldsymbol{q}_b}  
			+ \left.\nabla_{1}L_{\Delta} \right|_{\boldsymbol{q}_b, \boldsymbol{q}_c} \, , \\
			\boldsymbol{\mathcal{F}}_{\mathrm{D}} \left[\boldsymbol{F}_{\Delta}^{\pm}\right] (\boldsymbol{u}_a, \boldsymbol{u}_b, \boldsymbol{q}_a, \boldsymbol{q}_b, \boldsymbol{q}_c) 
			&\triangleq 
			\boldsymbol{F}_{\Delta}^+(\boldsymbol{u}_a, \boldsymbol{q}_a, \boldsymbol{q}_b) 
			+ \boldsymbol{F}_{\Delta}^-(\boldsymbol{u}_b, \boldsymbol{q}_b, \boldsymbol{q}_c) \, .
		\end{align}
	\end{subequations}
	Using these definitions, the discrete forced Euler-Lagrange equations \eqref{eq:DEL} at time step $n$ read
	\begin{equation}\label{eq:DEL_operator_form}
		\boldsymbol{\mathcal{L}}_{\mathrm{D}}\left[L_{\Delta}\right](\boldsymbol{q}_{n-1},{\boldsymbol{q}}_{n}, {\boldsymbol{q}}_{n+1}) 
		+ \boldsymbol{\mathcal{F}}_{\mathrm{D}} \left[\boldsymbol{F}_{\Delta}^{\pm}\right](\boldsymbol{u}_{n-1}, \boldsymbol{u}_{n}, \boldsymbol{q}_{n-1}, \boldsymbol{q}_n, \boldsymbol{q}_{n+1})  
		= \boldsymbol{0} \, .
	\end{equation}
	For conditioning the \ac{gp} priors \eqref{eq:dlgp_priors} on the discrete forced Euler-Lagrange equations \eqref{eq:DEL_operator_form}, it is straightforward to model the residual dynamics as an additive prior using the linear operators\footnote{Details on the construction and notation of operator-induced covariance functions are given in Appendix\,\ref{app:covariance_functions}.} \eqref{eq:dlgp_lin_operators}, \ie 
	\begin{align*}
		\boldsymbol{\mathcal{L}}_{\mathrm{D}}\left[L_{\Delta}\right] + \boldsymbol{\mathcal{F}}_{\mathrm{D}}\left[\boldsymbol{F}_{\Delta}^{\pm}\right] &\sim \mathcal{G} \mathcal{P}\left(\boldsymbol{0}, \boldsymbol{\mathcal{L}}_{\mathrm{D}} {\kappa}_L\left(\boldsymbol{r}, \boldsymbol{r}^{\prime}\right) \boldsymbol{\mathcal{L}}_{\mathrm{D}}^{\prime} + \boldsymbol{\mathcal{F}}_{\mathrm{D}}\boldsymbol{\kappa}_{F} \left(\boldsymbol{s}, \boldsymbol{s}^{\prime}\right)\boldsymbol{\mathcal{F}}_{\mathrm{D}}^{\prime} \right) \, .
	\end{align*}
	However, the resulting posterior yields trivial solutions, as already a degenerate null-Lagrangian satisfies the conditioning equation \eqref{eq:DEL_operator_form}, which features a \textcolor{red}{zero right-hand-side $\boldsymbol{y}=\boldsymbol{0}$}, $\boldsymbol{y}\in\mathbb{R}^{n_q}$, \ie
	\begin{equation*}
		\left\{ L_{\Delta} \mid  \boldsymbol{\mathcal{L}}_{\mathrm{D}}\left[L_{\Delta}\right] + \boldsymbol{\mathcal{F}}_{\mathrm{D}}\left[\boldsymbol{F}_{\Delta}^{\pm}\right]  + \boldsymbol{\epsilon}= \textcolor{red}{\boldsymbol{0}} \right\} \sim \mathcal{GP}\left( {m}^{L_{\Delta} | \textcolor{red}{\boldsymbol{0}}} , {{\kappa}}^{L_{\Delta} | \textcolor{red}{\boldsymbol{0}}} \right) \, ,
	\end{equation*} 
	where $\boldsymbol{\epsilon} \sim \mathcal{N}(\boldsymbol{0}, \boldsymbol{\Sigma})$, $\boldsymbol{\epsilon} \in \mathbb{R}^{n_q}$ is a noise term to introduce some slack \cite[Theorem\,1]{Pfortner.2022}. 
	Instead, a normalization condition is required to ensure that the learned Lagrangian is non-degenerate. 
	
	\paragraph{Normalization.} 
	Identifying a system's Lagrangian from trajectory observations is an ill-posed problem \cite{Offen.2025}, as the Lagrangian of a given system is non-unique. 
	In other words, a single observation trajectory can be explained by different Lagrangians. 
	Even more severe, the learned Lagrangian can be degenerate (\ie irregular), for instance, when a null-Lagrangian is found. 
	For discrete Lagrangians to be non-degenerate (\ie regular), $\frac{\partial^2 L_{\Delta}}{\partial \boldsymbol{q}_b \partial \boldsymbol{q}_a}$ needs to be invertible everywhere \cite{OberBlobaum.2011,Offen.2025}. 
	
	In most literature, the ambiguity and regularity of learned Lagrangians are not considered or only implicitly enforced via specific assumptions, such as specific Lagrangian structures or availability of external force measurements \cite{Dai.2024,Evangelisti.2022b,Evangelisti.2024b,Giacomuzzo.2024b}. 
	Thus, in contrast to previous work on learning non-conservative dynamics, we introduce normalization conditions to ensure the non-degeneracy of the learned Lagrangians, following \cite{Offen.2025}. 
    Details on the regularity and ambiguity of the learned Lagrangian are given in Appendix\,\ref{app:ambiguity_of_Lagrangians}. 

	Following the lines of \cite{Offen.2025}, we consider $2 n_q +1$ normalization conditions. 
	To this end, we enforce---at some anchor points $\bar{\boldsymbol{r}}_L$ and $\bar{\boldsymbol{s}}_F$---a fixed value $n_L \neq 0$ of the discrete Lagrangian through the evaluation operator $\mathcal{E}_{\bar{\boldsymbol{r}}_L} \left[L_{\Delta} \right]  \triangleq L_{\Delta} (\bar{\boldsymbol{r}}_L)$ and a fixed momentum $\boldsymbol{n}_M \neq \boldsymbol{0}$, $\boldsymbol{n}_M  \in \mathbb{R}^{n_q}$ through the discrete momentum operator $\boldsymbol{\mathcal{M}}_{\bar{\boldsymbol{r}}_L}^{+} \left[L_{\Delta} \right]  \triangleq \left.\nabla_{2} L_{\Delta} \right|_{\bar{\boldsymbol{r}}_L}$. 
    If additional physics knowledge is employed in the kernel design, one can relax the normalization conditions (see Appendices\,\ref{sec:phys_kernel_structure}, \ref{app:ambiguity_of_Lagrangians}, and \ref{app:experimental_details}). 
	Moreover, the condition $\boldsymbol{F}_{\Delta}^{\pm}(\boldsymbol{0}, \boldsymbol{0}, \boldsymbol{0}) \stackrel{}{=} \boldsymbol{0} =: \boldsymbol{n}_F$, $ \boldsymbol{n}_F \in \mathbb{R}^{n_q}$, is imposed via the evaluation operator $\boldsymbol{\mathcal{E}}_{\bar{\boldsymbol{s}}_F}\left[ \boldsymbol{F}_{\Delta}^{\pm} \right] \triangleq \boldsymbol{F}_{\Delta}^{\pm} (\bar{\boldsymbol{s}}_F)$, which ensures vanishing external forces at rest if $\boldsymbol{u} = \boldsymbol{0}$.  
    We formulate these normalization conditions as additional linear operators which are appended---for a training data set of size $N$---to $N$ evaluations of \eqref{eq:DEL_operator_form}, \ie
	\begin{equation}\label{eq:augm_el_operators}
		\bar{\boldsymbol{\mathcal{L}}}_{\mathrm{D}} \triangleq \begin{bmatrix} {\boldsymbol{\mathcal{L}}}_{\mathrm{D}} \\ \vdots \\ {\boldsymbol{\mathcal{L}}}_{\mathrm{D}}  \\ {\boldsymbol{\mathcal{C}}}_{{L}} \\ {\boldsymbol{0}} 
		\end{bmatrix} \, ,  \quad
		\bar{\boldsymbol{\mathcal{F}}}_{\mathrm{D}} \triangleq \begin{bmatrix}
			{\boldsymbol{\mathcal{F}}}_{\mathrm{D}} \\ \vdots \\ {\boldsymbol{\mathcal{F}}}_{\mathrm{D}} \\ {\boldsymbol{0}}  \\ {\boldsymbol{\mathcal{C}}}_{{F}}
		\end{bmatrix} \, , \; \qquad 
        \begin{aligned}
		&\mathrm{with} \quad &{\boldsymbol{\mathcal{C}}}_{{L}}\left[L_{\Delta} \right] &\triangleq \begin{bmatrix} \mathcal{E}_{\bar{\boldsymbol{r}}_L} \left[L_{\Delta} \right] \\ \boldsymbol{\mathcal{M}}_{\bar{\boldsymbol{r}}_L}^{+}\left[L_{\Delta} \right]
		\end{bmatrix} \, ,  \\
		&\mathrm{and} \quad &{\boldsymbol{\mathcal{C}}}_{{F}}\left[ \boldsymbol{F}_{\Delta}^{\pm} \right] &\triangleq 
		\boldsymbol{\mathcal{E}}_{\bar{\boldsymbol{s}}_F} \left[ \boldsymbol{F}_{\Delta}^{\pm} \right] \, ,
        \end{aligned}
	\end{equation}
    with the pseudo-measurement vector $\bar{\boldsymbol{y}}^\top = \left[\boldsymbol{y}^\top, \dots, \boldsymbol{y}^\top, {{n}}_L ,  {\boldsymbol{n}}_M^\top , {\boldsymbol{n}}_F^\top\right] \in \mathbb{R}^{(N+2) n_q +1}$. 
	The resulting joint normal distribution with normalization conditions is
	\begin{align}
		&\begin{bmatrix}
			L_{\Delta}  \\ \boldsymbol{F}_{\Delta}^{\pm} \\ \bar{\boldsymbol{y}} 
		\end{bmatrix} \sim \mathcal{N} \left(\begin{bmatrix}
			{0} \\ \boldsymbol{0} \\ \boldsymbol{0}  
		\end{bmatrix}, \begin{bmatrix}
			{\kappa}_L(\boldsymbol{r},\boldsymbol{r}^{\prime}) & \boldsymbol{0} &  {\kappa}_L \bar{\boldsymbol{\mathcal{L}}}_{\mathrm{D}}^{\prime}  \\
			\boldsymbol{0} & \boldsymbol{\kappa}_{F} \left(\boldsymbol{s}, \boldsymbol{s}^{\prime}\right) & \boldsymbol{\kappa}_F \bar{\boldsymbol{\mathcal{F}}}_{\mathrm{D}}^{\prime}  \\
			\bar{\boldsymbol{\mathcal{L}}}_{\mathrm{D}} {\kappa}_L & \bar{\boldsymbol{\mathcal{F}}}_{\mathrm{D}} \boldsymbol{\kappa}_F & \bar{\boldsymbol{\Theta}}_{\mathrm{D}}  \\
		\end{bmatrix}\right)\, ,  \label{eq:discrete_normal_dist}
    \end{align}
    which ensures regularity of the learned Lagrangian \cite{Offen.2025}. 
    See Appendix\,\ref{app:derivation} for a definition of $\bar{\boldsymbol{\Theta}}_{\mathrm{D}}$.

	\paragraph{Posteriors.} 
	Given the augmented linear operators $\bar{\boldsymbol{\mathcal{L}}}_{\mathrm{D}}$ and $\bar{\boldsymbol{\mathcal{F}}}_{\mathrm{D}}$, the \ac{gp} priors \eqref{eq:dlgp_priors} can be conditioned on the discrete forced Euler-Lagrange equations \cite{Pfortner.2022}. 
	The marginal posterior of the learned discrete Lagrangian is
	\begin{equation}\label{eq:dlgp_posterior}
		\begin{aligned}
			& \left\{ L_{\Delta}  \mid  \bar{\boldsymbol{\mathcal{L}}}_{\mathrm{D}}\left[L_{\Delta} \right] + \bar{\boldsymbol{\mathcal{F}}}_{\mathrm{D}}\left[\boldsymbol{F}_{\Delta}^{\pm}\right] + \bar{\boldsymbol{\epsilon}} = \bar{\boldsymbol{y}} \right\}  \sim \mathcal{GP}\left( {m}^{L_{\Delta}  | \bar{\boldsymbol{y}}} , {{\kappa}}^{L_{\Delta}  | \bar{\boldsymbol{y}}} \right) \, , \\
			&\hspace{1cm} {m}^{L_{\Delta}  | \bar{\boldsymbol{y}}} (\boldsymbol{r})= \bar{\boldsymbol{\mathcal{L}}}_{\mathrm{D}}\left[{\kappa}_L\left(\boldsymbol{r},\cdot\right)\right]^{\top} \bar{\boldsymbol{\Theta}}_{\mathrm{D}}^{\dagger} \bar{\boldsymbol{y}} \, ,\\
			&\hspace{1cm} {{\kappa}}^{L_{\Delta}  | \bar{\boldsymbol{y}}} (\boldsymbol{r}_1,\boldsymbol{r}_2) = {\kappa}_L\left( \boldsymbol{r}_1,\boldsymbol{r}_2 \right) -  \bar{\boldsymbol{\mathcal{L}}}_{\mathrm{D}}\left[{\kappa}_L\left(\boldsymbol{r}_1,\cdot\right)\right]^{\top} \bar{\boldsymbol{\Theta}}_{\mathrm{D}}^{\dagger} \bar{\boldsymbol{\mathcal{L}}}_{\mathrm{D}}\left[{\kappa}_L\left(\cdot,\boldsymbol{r}_2\right)\right]\, ,
		\end{aligned}
	\end{equation}
    with $\bar{\boldsymbol{\epsilon}}^\top = [{\boldsymbol{\epsilon}}^\top, \dots, {\boldsymbol{\epsilon}}^\top, \boldsymbol{0} ] \in \mathbb{R}^{(N+2) n_q +1}$. The marginal posterior of the learned discrete force is obtained analogously in Appendix\,\ref{app:derivation}.

	\subsection{Method 2: Learning Continuous Lagrangians and Forces}\label{sec:ct_lgp}
	Let us now introduce a continuous \ac{lgp} scheme for learning the continuous Lagrangian $L$ and forces $\boldsymbol{F}$ from discrete position data $\boldsymbol{q}$, leveraging a variational discretization method. 

	\paragraph{Priors.}
	We model the continuous Lagrangian $L$ and external forces $\boldsymbol{F}$ with zero-mean \acp{gp}
	\begin{align}\label{eq:clgp_priors}
		L  &\sim \mathcal{G} \mathcal{P}\left(0, k_L\left(\boldsymbol{z}, \boldsymbol{z}^{\prime}\right)\right) \, , \qquad \qquad \boldsymbol{F} \sim \mathcal{G} \mathcal{P}\left(\boldsymbol{0}, \boldsymbol{k}_{F} \left(\boldsymbol{x}, \boldsymbol{x}^{\prime}\right) \right) \, ,
	\end{align}
	where the variables $\boldsymbol{z}_n = \left(\boldsymbol{q}_n, \dot{\boldsymbol{q}}_n\right)$ and $\boldsymbol{x}_n = \left(\boldsymbol{u}_n, \boldsymbol{q}_n, \dot{\boldsymbol{q}}_n\right)$ summarize the arguments in \eqref{eq:contin_euler_lagrange}. 
	Analogously to Section\,\ref{sec:dt_lgp}, we model the multi-output force \ac{gp} as parallel single-output \acp{gp}.

	\paragraph{Variational discretization.} 
	To comply with \eqref{eq:DEL}, the continuous Lagrangian $L$ and force $\boldsymbol{F}$ must be discretized. 
	For this, we use variational discretization schemes that are linear in $L $ and $\boldsymbol{F}$, respectively, and derive linear operators to be incorporated in the previous operators $\boldsymbol{\mathcal{L}}_{\mathrm{D}}$ and $\boldsymbol{\mathcal{F}}_{\mathrm{D}}$. 
	This way, we \textit{by construction} integrate (discretize) all data entering (leaving) the \ac{gp} through the linear operators during training (prediction). 
	This contrasts with \cite{OberBlobaum.2023}, where only the predictive posterior employs variational discretization, while velocity data is required for \ac{lgp} training. 
	
	While any variational discretization scheme that is linear in $L $ and $\boldsymbol{F}$ can be employed, we use a midpoint rule with time step size $h$ and define the discretization operators 
	\begin{subequations}\label{eq:clgp_integrators}
		\begin{align}
			\boldsymbol{\mathcal{I}}_L \left[L \right](\boldsymbol{q}_a, \boldsymbol{q}_b) &\triangleq h L \left(\frac{\boldsymbol{q}_a + \boldsymbol{q}_b}{2}, \frac{\boldsymbol{q}_b - \boldsymbol{q}_a}{h}\right)\,  \quad &&\approx L_{\Delta}(\boldsymbol{q}_a, \boldsymbol{q}_b)  \, , \\
			\boldsymbol{\mathcal{I}}_F \left[\boldsymbol{F}\right](\boldsymbol{u}_a,\boldsymbol{q}_a, \boldsymbol{q}_b) & \triangleq \frac{h}{2} \boldsymbol{F} \left({\boldsymbol{u}_a}, \frac{\boldsymbol{q}_a + \boldsymbol{q}_b}{2}, \frac{\boldsymbol{q}_b - \boldsymbol{q}_a}{h}\right)\,  \quad && \approx \boldsymbol{F}_{\Delta}^{\pm}(\boldsymbol{u}_a, \boldsymbol{q}_a, \boldsymbol{q}_b)  \, .
		\end{align}
	\end{subequations}
	In contrast to many discrete dynamics models, the variational discretization \eqref{eq:clgp_integrators} enables prediction at custom prediction step sizes. 
	This is done by keeping  $h = \Delta t_{\mathrm{train}}$ for the training data and choosing $h = \Delta t_{\mathrm{pred}} $ with $\Delta t_{\mathrm{pred}} \neq \Delta t_{\mathrm{train}}$ for prediction at $(\boldsymbol{u}_a, \boldsymbol{q}_a, \boldsymbol{q}_b)$.

	\paragraph{Linear operators.} 
	To condition the continuous priors \eqref{eq:clgp_priors} on discrete forced Euler-Lagrange equations, we incorporate the variational discretization scheme in the linear operators \eqref{eq:dlgp_lin_operators} and define
	\begin{equation}\label{eq:clgp_operators}
		\begin{aligned}
			\boldsymbol{\mathcal{L}}_{\mathrm{DI}}\left[L \right](\boldsymbol{q}_{a},{\boldsymbol{q}}_{b}, {\boldsymbol{q}}_{c})  &\triangleq \nabla_{2} \boldsymbol{\mathcal{I}}_L \left[L \right](\boldsymbol{q}_{a}, \boldsymbol{q}_{b})  + \nabla_{1} \boldsymbol{\mathcal{I}}_L \left[L \right](\boldsymbol{q}_{b}, \boldsymbol{q}_{c}) \, , \\
			\boldsymbol{\mathcal{F}}_{\mathrm{DI}} \left[\boldsymbol{F}\right] (\boldsymbol{u}_{a}, \boldsymbol{u}_{b}, \boldsymbol{q}_{a}, \boldsymbol{q}_b, \boldsymbol{q}_{c})  &\triangleq \boldsymbol{\mathcal{I}}_F \left[\boldsymbol{F}\right](\boldsymbol{u}_{a},  \boldsymbol{q}_{a}, \boldsymbol{q}_{b}) + \boldsymbol{\mathcal{I}}_F \left[\boldsymbol{F}\right](\boldsymbol{u}_{b},   \boldsymbol{q}_b, \boldsymbol{q}_{c})  \, .
		\end{aligned}
	\end{equation}

	Using these definitions, the discrete forced Euler-Lagrange equations \eqref{eq:DEL} at time step $n$ can be reformulated as $\boldsymbol{\mathcal{L}}_{\mathrm{DI}}\left[L \right](\boldsymbol{q}_{n-1},{\boldsymbol{q}}_{n}, {\boldsymbol{q}}_{n+1}) + \boldsymbol{\mathcal{F}}_{\mathrm{DI}} \left[\boldsymbol{F}\right](\boldsymbol{u}_{n-1}, \boldsymbol{u}_{n},   \boldsymbol{q}_{n-1}, \boldsymbol{q}_n, \boldsymbol{q}_{n+1})  = \boldsymbol{0}$.
	
	\paragraph{Normalization and posteriors.} 
	The normalization conditions for continuous Lagrangians---required to ensure learning of non-degenerate Lagrangians---are similar to those introduced in Section\,\ref{sec:dt_lgp} \cite{Offen.2025}. 
	The resulting augmented linear operators $\bar{\boldsymbol{\mathcal{L}}}_{\mathrm{DI}}$ and $\bar{\boldsymbol{\mathcal{F}}}_{\mathrm{DI}}$ are then used to condition the \ac{gp} priors \eqref{eq:clgp_priors} \cite{Pfortner.2022}. 
	The marginal posteriors are defined analogously to those in Section \ref{sec:dt_lgp}, and the full derivation is given in Appendix\,\ref{app:derivation}.
	
	\conclusion{\textbf{In summary}, the discrete and continuous \acp{lgp} preserve the geometric structure of the Lagrange-d'Alembert principle by construction of the linear operators and, thereby, enable prediction without erroneous energy drift. 
		Specifically, the continuous scheme allows for prediction at custom step sizes by embedding a variational discretization. 
		Exploiting discrete forced Euler-Lagrange equations, our approach learns only from position snapshots $\boldsymbol{q}$, without requiring momentum or velocity measurements $\dot{\boldsymbol{q}}$. 
		The full methodological details are given in Appendix\,\ref{app:methods}, which includes approaches for incorporating further physics knowledge about $L$ and $\boldsymbol{F}$. 
		Equipped with these properties, the proposed \acp{lgp} are well-suited to provide stable, physically consistent long-term predictions in complex practical applications, which we test in the following.}

	\section{Experiments}\label{sec:results}
	
	We evaluate the performance of the \acp{lgp} in multiple synthetic and real-world case studies, including a pneumatically controlled real-world soft robot. 
	To this end, we compare continuous and discrete \acp{lgp} with and without additional energy model information. 
    We stress that we do not employ system-specific priors in the experimental evaluation, but rely only on the discrete forced Euler-Lagrange equations and---for some evaluations---a broadly applicable quadratic energy structure in the \ac{gp} kernels (see Appendix\,\ref{sec:phys_kernel_structure}). 
    Details on the employed models, experimental setups, resources, and further results are given in the Appendices\,\ref{app:experimental_details}--\ref{app:miscellanea}. 
	To the best of our knowledge, there is only one work \cite{Hansen.2025} that enables learning Lagrangians $L$ and external forces $\boldsymbol{F}$ solely from position data. 
	As their source code is not available, we compare the predictive performance of the \acp{lgp} with a standard \ac{gp} that provides one-step position predictions. 
	Multi-step predictions are generated by rollout from two known successive initial positions with a known input $\boldsymbol{u} : [t_0, t_N] \rightarrow U \subset \mathbb{R}^{n_u}$, $n_u = n_q$. 
	\ac{lgp} rollouts use a root finding algorithm that numerically solves \eqref{eq:DEL}. 
	The code will be available online\footnote{\texttt{https://github.com/link/to/be/added/for/final/conference/version}}.

	\subsection{Task 1: Controlled Multi-link Pendulum Simulation}\label{sec:results_sim_pendulum}
	We test the \acp{lgp} in a quantitative simulation study with controlled, damped multi-link pendulums. 
	We randomly generate $N \in \{25,50,100,200,300\}$ angle measurement triplets $\{\boldsymbol{q}_{n-1}^{(i)}, \boldsymbol{q}_{n}^{(i)}, \boldsymbol{q}_{n+1}^{(i)}\}_{i=1}^{N}$ from single, double, and triple pendulums, \ie $n_q \in\{1,2,3\}$, with varying torque inputs $\boldsymbol{u}$. 
	
\begin{figure}[tb]
		\centering
		
		
		\begin{subfigure}{0.58\textwidth}
			\centering
			\includegraphics[height=4cm]{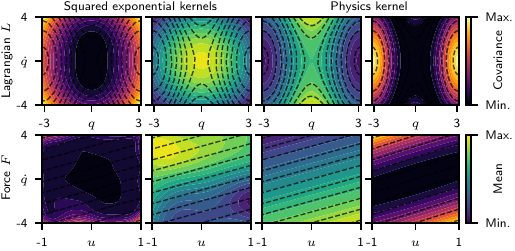}
			\caption{Learned Lagrangian and force \acsp{gp} in a single pendulum, \ie $n_q=1$ \\ ($N=300$ data points, dashed contours: true $L$ and $\boldsymbol{F}$).}
			\label{fig:pendulum_structure_Lagrangians}
		\end{subfigure}%
		\hfill
		\begin{subfigure}{0.4\textwidth}
			\centering
			\includegraphics[height=4cm]{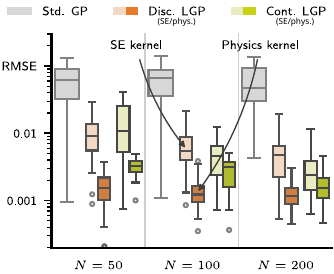}
			\caption{Error against training data budget in $20$-step simulations ($n_q=1$, $50$ random simulations).}
			\label{fig:pendulum_structure_prediction}
		\end{subfigure}%

		\caption{The proposed \acp{lgp} enable learning probabilistic models of a simulated pendulum's Lagrangian $L$ and external forces $\boldsymbol{F}$ only from position data. 
			Incorporating further physics knowledge into the kernel improves the accuracy of $L$ and $\boldsymbol{F}$ as well as the predictive performance. 
			See Appendices\,\ref{app:ambiguity_of_Lagrangians} and \ref{app:additional_results} for further details.
		}
        \vspace{-2mm}
		\label{fig:pendulum_structure}
	\end{figure}

	Figure\,\ref{fig:pendulum_structure_Lagrangians} illustrates how additional physics information in the kernel choice can improve the accuracy of the learned Lagrangian and force \acp{gp} in terms of their posterior mean and covariance. 
	We highlight that---without further assumptions---the Lagrangian and external force function of a system cannot be determined uniquely from trajectory data \cite{Offen.2025}. 
	Instead, a single trajectory can be generated from different \textit{equivalent} or \textit{alternative} Lagrangians \cite{Offen.2025}. 
	In this light, the \acp{lgp} learn the true Lagrangian up to a scaling operation, under certain conditions detailed in Appendix\,\ref{app:ambiguity_of_Lagrangians}. 
	Regardless, the \acp{lgp} provide a valid Euler-Lagrange operator, useful for prediction tasks. 
	A quantitative study of such predictions with varying system dimensions $n_q$ and varying amounts of training data $N$ shows that the \acp{lgp} outperform the standard \ac{gp} (see Figure\,\ref{fig:pendulum_structure_prediction} and Appendix\,\ref{app:case_studies_pendulum}). 
    Moreover, incorporating a physics-inspired kernel in the \acp{lgp} improves the predictive performance and the generalization beyond the training domain further (see Appendix\,\ref{app:ambiguity_of_Lagrangians}).

\begin{figure}[tb]
		\centering
		
		
		\begin{subfigure}{0.48\textwidth}
			\centering
			\includegraphics[height=3cm]{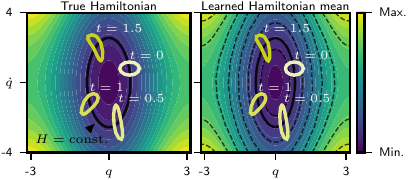}
			\caption{True and learned Hamiltonian (linear observable of $L$) and simulations for a set of initial conditions (green circles) ($n_q=1$, $N=300$, dashed contours: true $H$).}
			\label{fig:pendulum_2_structure_preservation}
		\end{subfigure}%
		\hfill
		\begin{subfigure}{0.48\textwidth}
			\centering
			\includegraphics[height=3cm]{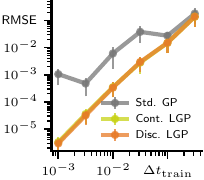}
			\includegraphics[height=3cm]{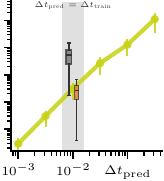}
			\caption{Error against training and prediction time step size in $20$-step simulations. Only cont. \ac{lgp} generalizes to $\Delta t_{\mathrm{pred}} \neq \Delta t_{\mathrm{train}}$ ($n_q=1$, $N=200$, $50$ random simulations).}
			\label{fig:pendulum_2_time_steps}
		\end{subfigure}%
		
		\caption{The proposed \acsp{lgp} yield structure-preserving and accurate long-term forward simulations of a controlled pendulum. Continuous \acsp{lgp} generalize to prediction time step sizes not seen during training $\Delta t_{\mathrm{pred}} \neq \Delta t_{\mathrm{train}}$.}
        \vspace{-2mm}
		\label{fig:pendulum_2}
	\end{figure}

	By using variational discretization schemes in Section\,\ref{sec:methods}, the learned \acp{lgp} preserve the symplectic structure of the Lagrange-d'Alembert principle \eqref{eq:lagrange_dalembert} by construction. 
	In practice, this means that a set of initial conditions preserves its volume if predicted forward in time. 
	In turn, this provides---in a conservative system, \ie without external forces---an approximately constant energy level of the forward predictions. 
	In fact, it oscillates around the initial energy, corresponding to constant-volume deformations of the set of initial conditions. 
	We observe this property in Figure\,\ref{fig:pendulum_2_structure_preservation} and Appendix\,\ref{app:pendulum_long_term_pred}, where we validate in a conservative pendulum that the structure-preserving \acp{lgp} yield accurate long-term predictions without energy drift, unlike the standard \ac{gp}. 
	As an aside, Figure\,\ref{fig:pendulum_2_structure_preservation} visualizes that we can compute any linear observable of the Lagrangian, such as the Hamiltonian, induced by the linear operator \eqref{eq:hamiltonian_legendre} 
	(see details in Appendix\,\ref{app:linear_observables}).
	In Figure\,\ref{fig:pendulum_2_time_steps}, we investigate the dependence on time step sizes. 
	First, we train at different time step sizes $\Delta t_{\mathrm{train}}$ and predict at time steps equal to those in the corresponding training data. 
	Second, we show that the continous-time \ac{lgp} generalizes to custom prediction steps, \ie $\Delta t_{\mathrm{train}} \neq \Delta t_{\mathrm{pred}}$, thanks to the embedded discretization operator (see Figure\,\ref{fig:pendulum_2_time_steps} for $\Delta t_{\mathrm{train}} = 10^{-2}$).

	\subsection{Task 2: Controlled Real-World Double Pendulum}\label{sec:results_rw_pendulum}
\begin{figure}[tb]
		\centering
		\sbox{\leftcolboxa}{%
			\begin{minipage}[c]{0.32\textwidth}
				
				\begin{minipage}{\textwidth}
					\centering
					\includegraphics[width=\linewidth]{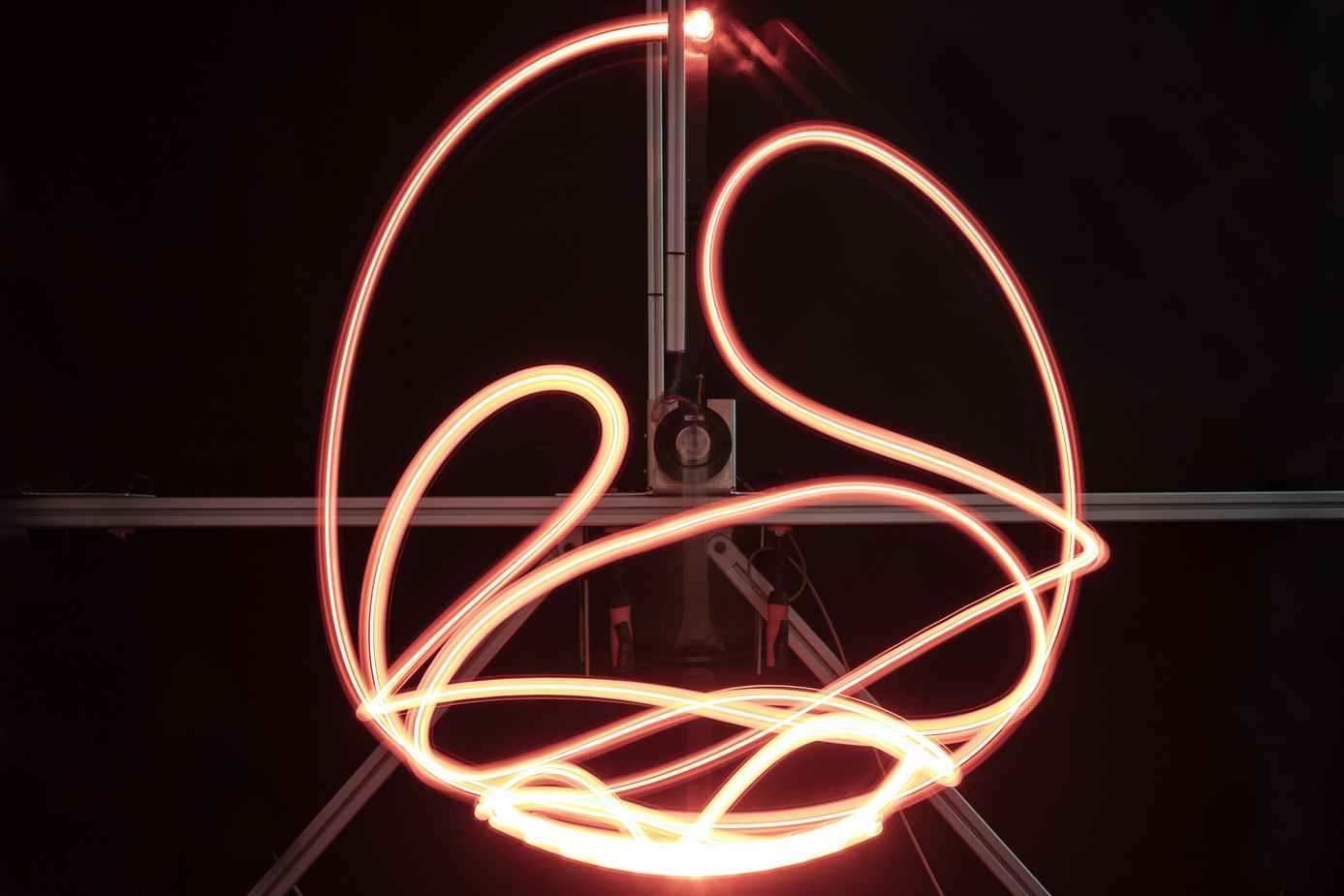}
				\end{minipage}
				
				\vspace{0.1cm} 
				
				\begin{minipage}{\textwidth}
					\centering
					{\fontsize{8pt}{8pt}\selectfont
						\resizebox{1\linewidth}{!}{
\begin{tabular}{lrrrr}
\toprule[1pt]
& & & \multiline{\hyperref[fig:real_world_pendulum_app]{\textbf{\acs{rmse} Task 2.1}}} & \multiline{\hyperref[fig:real_world_pendulum]{\textbf{\acs{rmse} Task 2.2}} $\rightarrow$} \\
Method & $\Delta t_{\mathrm{pred}}$ &  & \multiline{with input} & \multiline{dissipation only} \\
\cmidrule{1-2} \cmidrule{4-5}
\textbf{\textcolor[HTML]{787878}{Std. \acs{gp}}} & \textcolor[HTML]{787878}{$61\,\mathrm{ms}$} & & \textcolor[HTML]{787878}{$0.49\,\mathrm{rad}$} & \textcolor[HTML]{787878}{$0.54\,\mathrm{rad}$} \\
\textbf{\textcolor[HTML]{E77B29}{Disc. \acs{lgp}}} & \textcolor[HTML]{E77B29}{$61\,\mathrm{ms}$} & & \textcolor[HTML]{E77B29}{$0.12\,\mathrm{rad}$} & \textcolor[HTML]{E77B29}{$0.17\,\mathrm{rad}$} \\
\textbf{\textcolor[HTML]{C8D317}{Cont. \acs{lgp}}} & \textcolor[HTML]{C8D317}{$61\,\mathrm{ms}$} & & \textcolor[HTML]{C8D317}{$0.17\,\mathrm{rad}$} & \textcolor[HTML]{C8D317}{$0.23\,\mathrm{rad}$} \\
\textbf{\textcolor[HTML]{00509B}{Cont. \acs{lgp}}} & \textcolor[HTML]{00509B}{$100\,\mathrm{ms}$} & & \textcolor[HTML]{00509B}{$0.23\,\mathrm{rad}$} & \textcolor[HTML]{00509B}{$0.18\,\mathrm{rad}$} \\
\bottomrule[1pt]
\end{tabular}
}}
				\end{minipage}
				
			\end{minipage}%
		}
		\usebox{\leftcolboxa}%
		\hfill 
		\begin{minipage}[c]{0.64\textwidth}
			\centering
			\includegraphics[height=\dimexpr\ht\leftcolboxa+\dp\leftcolboxa\relax, width=\linewidth]{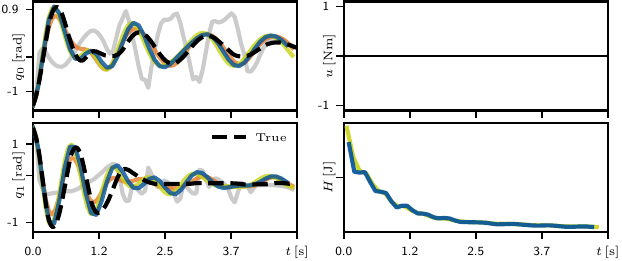}
		\end{minipage}
		
		\caption{Prediction tasks in a controlled real-world double pendulum. The \acsp{lgp} yield accurate forward simulations despite learning only from noisy real-world position data ($\Delta t_{\mathrm{train}}=61\,\mathrm{ms}$, $n_q=2$, $N=300$ data points). In the absence of inputs, the learning-based system energy (Hamiltonian $H$) along the trajectory---consistent with the underlying physics---decays due to dissipation. Task 2.1 is visualized in Figure\,\ref{fig:real_world_pendulum_app}. Photo and data adopted from GitHub-Repository of \cite{Wiebe.2024}.}
        \vspace{-2mm}
	\label{fig:real_world_pendulum}
\end{figure}

	Second, we evaluate the prediction performance with a real-world double pendulum data set \cite{Kumar.2025}. 
	The pendulum features angular positions $\boldsymbol{q}$ and is non-conservative in two ways. 
	First, the joints are driven by a torque input $\boldsymbol{u}$, and second, the pendulum is subject to friction, which dissipates energy over time. 
    For training, we only use $N=300$ position and input torque data triples, randomly sampled from the noisy trajectory measurements.
	For testing, we choose two $5\,\mathrm{s}$-prediction scenarios, one with non-zero motor torque inputs $\boldsymbol{u}$ and one without inputs, \ie dissipation only. 
	The predictions in Figures\,\ref{fig:real_world_pendulum} and \ref{fig:real_world_pendulum_app} underscore the capability of the structure-preserving \acp{lgp} to perform accurate long-term predictions, in contrast to the baseline \ac{gp}. 
	Importantly, the learned \acp{lgp} yield a monotonically decreasing system energy (Hamiltonian $H$) in the absence of control inputs, which is consistent with the dissipative nature of the pendulum. 
	We stress that the continuous \ac{lgp} enables stable predictions at \textit{larger time step sizes}, \ie $\Delta t_{\mathrm{train}} < \Delta t_{\mathrm{pred}}$, while staying \textit{on par} regarding the prediction error. 
	
\subsection{Task 3: Controlled Real-World Soft Robot}\label{sec:results_rw_soft_robot}
\begin{figure}[tb]
	\centering
	\sbox{\leftcolboxb}{%
		\begin{minipage}[c]{0.23\textwidth}
			
			\begin{minipage}{\textwidth}
				\centering
				\includegraphics[width=\linewidth]{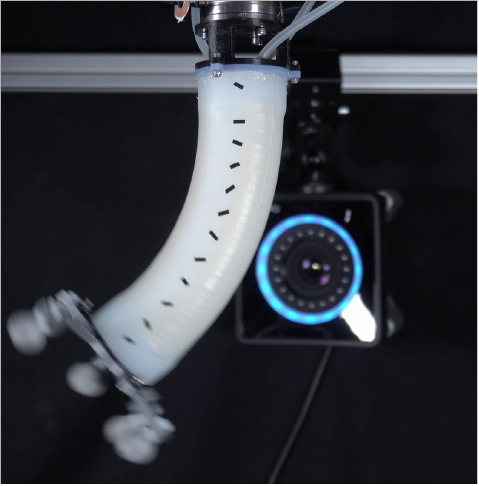}
			\end{minipage}
			
			\vspace{0.1cm} 
			
			\begin{minipage}{\textwidth}
				\centering
				{\fontsize{8pt}{8pt}\selectfont
					\resizebox{1\linewidth}{!}{
\begin{tabular}{lrrr}
\toprule[1pt]
 &  & & \multiline{\textbf{\acs{rmse} Task 3}} \\
Method & $\Delta t_{\mathrm{pred}}$ &  & \multiline{input \& dissip.} \\
\cmidrule{1-2} \cmidrule{4-4}
\textbf{\textcolor[HTML]{787878}{Std. \acs{gp}}} & \textcolor[HTML]{787878}{$20\,\mathrm{ms}$} & & \textcolor[HTML]{787878}{$3.37\,\mathrm{mm}$} \\
\textbf{\textcolor[HTML]{E77B29}{Disc. \acs{lgp}}} & \textcolor[HTML]{E77B29}{$20\,\mathrm{ms}$} & & \textcolor[HTML]{E77B29}{$0.64\,\mathrm{mm}$} \\
\textbf{\textcolor[HTML]{C8D317}{Cont. \acs{lgp}}} & \textcolor[HTML]{C8D317}{$20\,\mathrm{ms}$} & & \textcolor[HTML]{C8D317}{$0.57\,\mathrm{mm}$} \\
\textbf{\textcolor[HTML]{00509B}{Cont. \acs{lgp}}} & \textcolor[HTML]{00509B}{$40\,\mathrm{ms}$} & & \textcolor[HTML]{00509B}{$0.54\,\mathrm{mm}$} \\
\bottomrule[1pt]
\end{tabular}
}}
	\end{minipage}
	
\end{minipage}%
}

\usebox{\leftcolboxb}%
\hfill 
\begin{minipage}[c]{0.75\textwidth}
\centering

\includegraphics[height=\dimexpr\ht\leftcolboxb+\dp\leftcolboxb\relax, width=\linewidth]{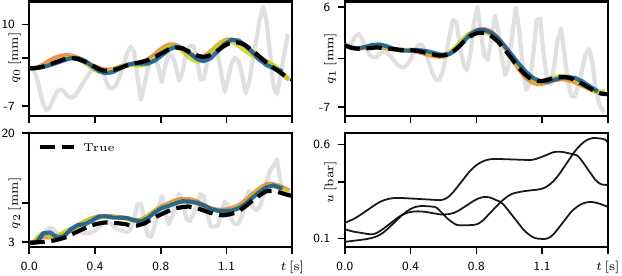}
\end{minipage}

\caption{Prediction task in a controlled pneumatic real-world soft robot. Although the soft robot exhibits complex nonlinear kinematics and dynamics with hysteresis effects, the \acsp{lgp} enable accurate forward simulations of the shape-describing parameters $\boldsymbol{q}^\top = [\Delta x, \Delta y, \delta \ell]$ ($\Delta t_{\mathrm{train}}=20\,\mathrm{ms}$, $n_q=3$, $N=200$ data points). Photo and data adopted from \cite{Mehl.2024}.}
\vspace{-2mm}
\label{fig:real_soft_robot}
\end{figure}

The previous test scenarios featured comparably obvious generalized coordinates, \ie the positions $\boldsymbol{q}$, that clearly describe the dynamics to be learned. 
In contrast, the controlled real-world soft robot in Figure\,\ref{fig:real_soft_robot} exhibits---besides highly nonlinear dynamics---no straightforward ``position coordinates.'' 
Instead, we perform training and prediction based on so-called ``shape-parameters'' $\boldsymbol{q}^\top = [\Delta x, \Delta y, \delta \ell]$ \cite{DellaSantina.2020}, obtained via a motion tracking set-up \cite{Mehl.2024}. 
Notably, the robot's silicon structure is non-rigid and is driven by three pneumatic pressure inputs $\boldsymbol{u}$, which together yield complex, non-conservative dynamics with hysteresis. 
Despite the challenging system behavior and the unknown system dimension, the \acp{lgp} provide accurate long-term predictions, unlike the standard \ac{gp}. 

\section{Related Work}\label{sec:rel_works}

\paragraph{Energy-consistent dynamics learning.} Integrating continuous-time dynamics into learning-based models is a popular and related approach. 
Early work in this direction learns vector-valued flow maps that determine the evolution of a system's state, \eg using neural ODEs \cite{Chen.2018}, but does not incorporate stronger inductive physics biases. 
In contrast, various recent work computes a flow map from an approximation of a system's Hamiltonian \cite{Beckers.2022,Greydanus.2019,Tanaka.2022,Sosanya.2022,Roth.2025,Desai.2021,Bertalan.2019,Rath.2021,Offen.2022,Ensinger.2023,Ross.2023,Ross.2024,Hu.2025,Ewering.2025,Ensinger.2023} or Lagrangian \cite{Xiao.2024,Trinh.2025,Cranmer.2019,Lutter.2019b,Giacomuzzo.2024,Giacomuzzo.2024b,OberBlobaum.2023,Offen.2025,Evangelisti.2022b,Dai.2024}, which reflect the exchange of energy within the system. 
Most of these works apply continuous Euler-Lagrange or Hamiltonian mechanics equations---induced by the Lagrange-d'Alembert principle \eqref{eq:lagrange_dalembert}---to the approximated Hamiltonian/Lagrangian for constructing the flow map \cite{Xiao.2024,Trinh.2025,Sosanya.2022,Roth.2025,Desai.2021,Cranmer.2019,Lutter.2019b,Greydanus.2019,Beckers.2022,Tanaka.2022,Bertalan.2019,Rath.2021,Ensinger.2023,Ross.2023,Ross.2024,Hu.2025,Ewering.2025,Giacomuzzo.2024,Giacomuzzo.2024b,Evangelisti.2022b,Dai.2024}. 
This usually requires access to momentum or velocity measurements, which is a restrictive assumption in many practical settings \cite{Hansen.2025}. 
While velocity and momentum estimates can be approximated from position measurements via numerical differentiation, this approach often introduces significant inaccuracies due to noise \cite{Chartrand.2011}. 
In contrast, we circumvent these issues via discrete Euler-Lagrange equations \eqref{eq:DEL}, which only require positional data \cite{OberBlobaum.2023,Offen.2025,Hansen.2025,Lishkova.2023,Duruisseaux.2023}. 
In addition, existing energy-consistent learning methods---unlike the proposed schemes---often rely on strong assumptions about the kernel design, such as a quadratic kinetic energy, to ensure regularity of the learned Lagrangian \cite{Evangelisti.2022b,Giacomuzzo.2024}. 

\paragraph{Structure preservation of energy principles.} When simulating dynamics from a learned flow map, special attention is required regarding the underlying symplectic structure. 
Specifically, for a dynamical system with system energy (Hamiltonian) $H$, its variational symmetries yield conserved quantities of the flow, such as energy conservation, by Noether’s theorem. 
Various works require symplectic integration schemes to preserve this structure, as they utilize continuous Euler-Lagrange equations, or even ignore structure preservation altogether, causing the learned model to violate the Lagrange-d'Alembert principle \cite{Cranmer.2019,Lutter.2019b,Giacomuzzo.2024,Giacomuzzo.2024b,Evangelisti.2024b,Evangelisti.2022b,Dai.2024,Beckers.2022}. 
The proposed methods differ from most of these approaches by employing discrete forced Euler-Lagrange equations \eqref{eq:DEL} that preserve---in the absence of external forces---the symplectic structure of the underlying energy principle by construction \cite{Offen.2025,OberBlobaum.2023,Hansen.2025,Brudigam.2022}. 
Considering the continuous \ac{lgp} in Section\,\ref{sec:ct_lgp}, the symplectic structure is implicitly preserved as all data enters the \ac{gp} through a variational integrator, embedded in the linear operators $\boldsymbol{\mathcal{L}}_{\mathrm{DI}}$ and $\boldsymbol{\mathcal{F}}_{\mathrm{DI}}$. 
This procedure relates to \cite{OberBlobaum.2023}, where a similar approach is applied to the \ac{gp}'s posterior.

\paragraph{Learning dynamics with dissipation and inputs.} Seminal works on Lagrangian or Hamiltonian-based models \cite{Cranmer.2019,Lutter.2019b,Greydanus.2019} focused on conservative dynamical systems, meaning that no energy crosses the system boundary, \eg via dissipation or inputs. 
Recent extensions enable learning with such forcing terms \cite{Dai.2024,Evangelisti.2024b,Xiao.2024,Hansen.2025,Beckers.2022,Roth.2025,Trinh.2025,Tanaka.2022, Li.2026}, enabling practical implementation in several practically relevant modeling problems, for instance, in robotics \cite{Liu.2024,Weiss.2026, Li.2026}. 
Considering existing approaches that enable learning from position data, most work \cite{Offen.2025,OberBlobaum.2023,Lishkova.2023} is restricted to conservative systems by building on the principle of least action, \ie the first term in \eqref{eq:lagrange_dalembert}.
Instead, the proposed \acp{lgp} rely on the \textit{entire} Lagrange-d'Alembert principle \eqref{eq:lagrange_dalembert}, which admits learning non-conservative dynamics.

\paragraph{Uncertainty quantification.} Considering methods for energy-based dynamics learning, a significant part of the literature focuses on deterministic neural networks, such as \acp{lnn} or \acp{hnn} \cite{Xiao.2024,Trinh.2025,Sosanya.2022,Roth.2025,Desai.2021,Cranmer.2019,Lutter.2019b,Greydanus.2019, Hansen.2025,Duruisseaux.2023,Li.2026}. 
In contrast, schemes providing an uncertainty-quantification are almost exclusively based on \acp{gp} \cite{Beckers.2022,Tanaka.2022,Bertalan.2019,Rath.2021,Ensinger.2023,Ross.2023,Ross.2024,Hu.2025,Ewering.2025,Giacomuzzo.2024,Giacomuzzo.2024b,OberBlobaum.2023,Offen.2025,Evangelisti.2022b,Dai.2024}. 
To the best of our knowledge, the proposed \acp{lgp} are the first methods to enable learning of non-conservative \textit{probabilistic} dynamics models from position data.

\section{Discussion}

\paragraph{Limitations.} The proposed \acp{lgp} inherit the typical limitations and properties of \acp{gp}, including their limited scalability with the number of training data points. 
Thus, future work may consider incorporating sparse \ac{gp} approaches.
Considering the required prior knowledge to model dynamics using \acp{lgp}, we acknowledge that it is necessary to assume a system dimension. 
Yet, we have shown in a real-world soft robot---whose actual spatial coordinates are continuous and thus infinite-dimensional---that the \acp{lgp} model performs well even if the true system dimension is not matched. 
Moreover, while we test on noisy real-world data, we are aware that the considered midpoint rule for variational discretization may be sensitive to higher levels of sensor noise. 
An effective ad hoc countermeasure is to raise the signal-to-noise ratio by increasing the training time step size. 
However, this comes at the cost of potentially losing high-frequency features. 
In this regard, future work may investigate the effect of other variational integration schemes. 

\paragraph{Conclusion.} In this paper, we propose \aclp{lgp} (\acsp{lgp}) for learning probabilistic, non-conservative dynamics models. 
Both presented schemes, discrete and continuous \ac{lgp}, preserve the geometric structure of the underlying Lagrange-d'Alembert principle by construction, in the unforced case. 
By conditioning on discrete forced Euler-Lagrange equations, the \acp{lgp} learn \textit{only from position data}. 
To the best of our knowledge, the proposed \acp{lgp} enable, for the first time, learning non-conservative dynamics from position data while providing an uncertainty quantification. 
While our approach enables learning with generic kernels, adding physics knowledge into kernel design improves data efficiency, generalization, and predictive performance. 
Various synthetic and real-world case studies---including a pneumatic real-world soft robot---show that the \acp{lgp} yield highly accurate, physically consistent long-term predictions in complex real-world applications. 

\begin{ack}
This research was partially supported by \emph{Kjell och M{\"a}rta Beijer Foundation} and by the projects \emph{Blending probabilistic and nonlinear representations} (contract number: 2025-04318) and \emph{Physics-informed machine learning} (contract number: 2021-04321), funded by the Swedish Research Council. 
Moreover, the research was partially supported by \textit{German Academic Scholarship Foundation (Studienstiftung des Deutschen Volkes)}. 
\end{ack}

\bibliographystyle{unsrtnat} 
\bibliography{neurips2026}      


\newpage
\appendix

    \setcounter{figure}{0}
    \setcounter{table}{0}
    \setcounter{equation}{0}
    \renewcommand{\thefigure}{A\arabic{figure}}
    \renewcommand{\theequation}{A\arabic{equation}}
    \renewcommand{\thetable}{A\arabic{table}}
    \renewcommand{\theHfigure}{A\arabic{figure}}
    \renewcommand{\theHtable}{A\arabic{table}}
    \newtheorem{proposition}{Proposition}
    \newtheorem*{proposition*}{Proposition}

    \onecolumn
    \pagenumbering{roman} 
    \mtcsettitle{parttoc}{} 
    \setcounter{secnumdepth}{-2} 
    \part{Appendix}              
    \parttoc
    \setcounter{secnumdepth}{3}  

    \setcounter{page}{0 }

\newpage
\section{Method Details}\label{app:methods}
	In this appendix, we give details on the construction of covariance functions (Appendix\,\ref{app:covariance_functions}), the full derivation of the \acp{lgp} (Appendix\,\ref{app:derivation}), and how to construct linear observables of the learned quantities (Appendix\,\ref{app:linear_observables}). 
    In Appendix\,\ref{sec:phys_kernel_structure}, we explain how additional physics knowledge can be incorporated in the kernel design. 
	Last, in Appendix\,\ref{app:ambiguity_of_Lagrangians}, we elaborate on the ambiguity of Lagrangians and its effect on dynamics learning and prediction.

\subsection{Construction of Covariance Functions}\label{app:covariance_functions}

We detail how the covariance functions are constructed from linear operators. 
To this end, we adopt the notation of \cite[Notation 1]{Pfortner.2022}.

\begin{ass}[Assumption 1 in \cite{Pfortner.2022}]
	Let $f\sim\mathcal{GP}(m,k)$ be a Gaussian process prior with index set $\mathbb{X}$ on the probability space $(\Omega, \mathcal{F}, \mathrm{P})$, whose paths lie in a real separable reproducing kernel Banach space (RKBS) $\mathbb{B} \subset \mathbb{R}^{\mathbb{X}}$ such that $\omega \mapsto f(\cdot, \omega)$ is a $\mathbb{B}$-valued Gaussian random variable.
\end{ass}

Let $f\sim\mathcal{GP}(m,k)$ satisfy Assumption~1, and let
\begin{equation}
\boldsymbol{\mathcal{L}}:\mathcal{B}\to\mathbb{R}^{n},
\tilde{\boldsymbol{\mathcal{L}}}:\mathcal{B}\to\mathbb{R}^{\tilde n}
\end{equation}
be bounded linear operators.
Given this and following \cite{Pfortner.2022}, define the entries of the matrix $\boldsymbol{\mathcal{L}} k\tilde{\boldsymbol{\mathcal{L}}}' \in \mathbb{R}^{n \times \tilde{n}}$ as
\begin{equation}
\left(\boldsymbol{\mathcal{L}}k\tilde{\boldsymbol{\mathcal{L}}}'\right)_{ij}
\;:=\;
\boldsymbol{\mathcal{L}}\!\left[\boldsymbol{x}\mapsto \tilde{\boldsymbol{\mathcal{L}}}\!\left[k(\boldsymbol{x},\cdot)\right]_j\right]_i .
\label{eq:notation1_entry}
\end{equation}
In this case, the order of application is interchangeable
\begin{equation}
\boldsymbol{\mathcal{L}}\!\left[\boldsymbol{x}\mapsto \tilde{\boldsymbol{\mathcal{L}}}\!\left[k(\boldsymbol{x},\cdot)\right]_j\right]_i
=
\tilde{\boldsymbol{\mathcal{L}}}\!\left[\boldsymbol{x}\mapsto \boldsymbol{\mathcal{L}}\!\left[k(\cdot,\boldsymbol{x})\right]_i\right]_j ,
\label{eq:notation1_commute}
\end{equation}
which motivates the parenthesis-free shorthand $\boldsymbol{\mathcal{L}} k\tilde{\boldsymbol{\mathcal{L}}}'$.

\subsection{Derivation Details}\label{app:derivation}

\subsubsection{Discrete Lagrangian Gaussian Processes}

Most of the derivation of the discrete \ac{lgp} scheme can be found in Section\,\ref{sec:dt_lgp} of the main paper. 
For completeness, we here give the joint normal distribution with normalization conditions \eqref{eq:discrete_normal_dist}, \ie
	\begin{align*}
		&\begin{bmatrix}
			L_{\Delta}  \\ \boldsymbol{F}_{\Delta}^{\pm} \\ \bar{\boldsymbol{y}} 
		\end{bmatrix} \sim \mathcal{N} \left(\begin{bmatrix}
			{0} \\ \boldsymbol{0} \\ \boldsymbol{0}  
		\end{bmatrix}, \begin{bmatrix}
			{\kappa}_L(\boldsymbol{r},\boldsymbol{r}^{\prime}) & \boldsymbol{0} &  {\kappa}_L \bar{\boldsymbol{\mathcal{L}}}_{\mathrm{D}}^{\prime}  \\
			\boldsymbol{0} & \boldsymbol{\kappa}_{F} \left(\boldsymbol{s}, \boldsymbol{s}^{\prime}\right) & \boldsymbol{\kappa}_F \bar{\boldsymbol{\mathcal{F}}}_{\mathrm{D}}^{\prime}  \\
			\bar{\boldsymbol{\mathcal{L}}}_{\mathrm{D}} {\kappa}_L & \bar{\boldsymbol{\mathcal{F}}}_{\mathrm{D}} \boldsymbol{\kappa}_F & \bar{\boldsymbol{\Theta}}_{\mathrm{D}}  \\
		\end{bmatrix}\right)\, ,  
    \end{align*}
where the matrix $\bar{\boldsymbol{\Theta}}_{\mathrm{D}} \in \mathbb{R}^{[(N+2)n_q +1] \times [(N+2)n_q +1]}$ is defined as 
\begin{align}
\bar{\boldsymbol{\Theta}}_{\mathrm{D}} \triangleq 
\left[
\begin{array}{cccc|cc}
    \boldsymbol{\Theta}_{\mathrm{D}, 11}+ \boldsymbol{\Sigma} & \boldsymbol{\Theta}_{\mathrm{D}, 12} & \dots & \boldsymbol{\Theta}_{\mathrm{D}, 1N} & \boldsymbol{\mathcal{L}}_{\mathrm{D},1} {\kappa}_L \boldsymbol{\mathcal{C}}_L^{\prime} & \boldsymbol{\mathcal{F}}_{\mathrm{D},1}\boldsymbol{\kappa}_{F}\boldsymbol{\mathcal{C}}_F^{\prime}\\
    \boldsymbol{\Theta}_{\mathrm{D}, 21} & \boldsymbol{\Theta}_{\mathrm{D},22}+ \boldsymbol{\Sigma} & \dots & \boldsymbol{\Theta}_{\mathrm{D}, 2N} & \boldsymbol{\mathcal{L}}_{\mathrm{D},2} {\kappa}_L \boldsymbol{\mathcal{C}}_L^{\prime} & \boldsymbol{\mathcal{F}}_{\mathrm{D},2}\boldsymbol{\kappa}_{F}\boldsymbol{\mathcal{C}}_F^{\prime}\\
    \vdots & \vdots & \ddots & \vdots & \vdots & \vdots \\
    \boldsymbol{\Theta}_{\mathrm{D}, N1} & \boldsymbol{\Theta}_{\mathrm{D}, N2} & \dots & \boldsymbol{\Theta}_{\mathrm{D},NN}+ \boldsymbol{\Sigma} & \boldsymbol{\mathcal{L}}_{\mathrm{D},N} {\kappa}_L \boldsymbol{\mathcal{C}}_L^{\prime} & \boldsymbol{\mathcal{F}}_{\mathrm{D},N}\boldsymbol{\kappa}_{F}\boldsymbol{\mathcal{C}}_F^{\prime}\\
    \hline
    \boldsymbol{\mathcal{C}}_L \kappa_L \boldsymbol{\mathcal{L}}_{\mathrm{D},1}^{\prime} & \boldsymbol{\mathcal{C}}_L \kappa_L \boldsymbol{\mathcal{L}}_{\mathrm{D},2}^{\prime} & \dots & \boldsymbol{\mathcal{C}}_L \kappa_L \boldsymbol{\mathcal{L}}_{\mathrm{D},N}^{\prime} & \boldsymbol{\mathcal{C}}_L \kappa_L \boldsymbol{\mathcal{C}}_L^{\prime} & \boldsymbol{0} \\
    \boldsymbol{\mathcal{C}}_F\boldsymbol{\kappa}_F\boldsymbol{\mathcal{F}}_{\mathrm{D},1}^{\prime}  & \boldsymbol{\mathcal{C}}_F\boldsymbol{\kappa}_F\boldsymbol{\mathcal{F}}_{\mathrm{D},2}^{\prime} & \dots & \boldsymbol{\mathcal{C}}_F\boldsymbol{\kappa}_F\boldsymbol{\mathcal{F}}_{\mathrm{D},N}^{\prime} & \boldsymbol{0} & \boldsymbol{\mathcal{C}}_F\boldsymbol{\kappa}_F \boldsymbol{\mathcal{C}}_F^{\prime}
\end{array}
\right]\, .
\label{eq:theta_matrix_expanded}
\end{align}
The matrices $\boldsymbol{\Theta}_{\mathrm{D},ij} \in \mathbb{R}^{n_q \times n_q}$ are the covariances of residual dynamics at two data points $i$ and $j$, \ie
\begin{equation}
\boldsymbol{\Theta}_{\mathrm{D},ij} \triangleq \boldsymbol{\mathcal{F}}_{\mathrm{D},i} \boldsymbol{\kappa}_F \boldsymbol{\mathcal{F}}_{\mathrm{D},j}^{\prime} + \boldsymbol{\mathcal{L}}_{\mathrm{D},i} {\kappa}_L \boldsymbol{\mathcal{L}}_{\mathrm{D},j}^{\prime} \, , \qquad \forall i, j \in [1,N] \,.
\end{equation}
Thus, we require $N$ data triplets $\{\boldsymbol{q}_{n-1}^{(i)}, \boldsymbol{q}_{n}^{(i)}, \boldsymbol{q}_{n+1}^{(i)}\}_{i=1}^{N}$ with corresponding inputs $\{ \boldsymbol{u}_{n-1}^{(i)},\boldsymbol{u}_{n}^{(i)} \}_{i=1}^{N} $ to construct $\bar{\boldsymbol{\Theta}}_{\mathrm{D}}$ in \eqref{eq:theta_matrix_expanded}.

\paragraph{Posteriors.} The resulting marginal posterior distribution of the discrete Lagrangian is given in \eqref{eq:dlgp_posterior}. 
 Analogously, we here give the discrete external forces $\boldsymbol{F}_{\Delta}^{\pm}$, conditioned on the discrete forced Euler-Lagrange operators \eqref{eq:DEL_operator_form}. 
 The marginal posterior
\begin{equation}
    	\begin{aligned}
        		& \left\{ \boldsymbol{F}_{\Delta}^{\pm}  \mid  \bar{\boldsymbol{\mathcal{L}}}_{\mathrm{D}}\left[L_{\Delta} \right] + \bar{\boldsymbol{\mathcal{F}}}_{\mathrm{D}}\left[\boldsymbol{F}_{\Delta}^{\pm}\right] + \bar{\boldsymbol{\epsilon}} = \bar{\boldsymbol{y}} \right\}  \sim \mathcal{GP}\left( \boldsymbol{m}^{\boldsymbol{F}_{\Delta}^{\pm}  | \bar{\boldsymbol{y}}} , \boldsymbol{\kappa}^{\boldsymbol{F}_{\Delta}^{\pm}  | \bar{\boldsymbol{y}}} \right) \, ,\\
        		& \hspace{1cm} \boldsymbol{m}^{\boldsymbol{F}_{\Delta}^{\pm} | \bar{\boldsymbol{y}}} (\boldsymbol{s})= \bar{\boldsymbol{\mathcal{F}}}_{\mathrm{D}}\left[\boldsymbol{\kappa}_F\left(\boldsymbol{s},\cdot\right)\right]^{\top} \bar{\boldsymbol{\Theta}}_{\mathrm{D}}^{\dagger} \bar{\boldsymbol{y}} \, ,\\
        		& \hspace{1cm} \boldsymbol{\kappa}^{\boldsymbol{F}_{\Delta}^{\pm} | \bar{\boldsymbol{y}}} (\boldsymbol{s}_1,\boldsymbol{s}_2)= \boldsymbol{\kappa}_F \left( \boldsymbol{s}_1,\boldsymbol{s}_2 \right)-  \bar{\boldsymbol{\mathcal{F}}}_{\mathrm{D}}\left[\boldsymbol{\kappa}_F\left(\boldsymbol{s}_1,\cdot\right)\right]^{\top} \bar{\boldsymbol{\Theta}}_{\mathrm{D}}^{\dagger} \bar{\boldsymbol{\mathcal{F}}}_{\mathrm{D}}\left[\boldsymbol{\kappa}_F\left(\cdot,\boldsymbol{s}_2\right)\right]\, ,
        	\end{aligned}
    \end{equation}
directly follows from the joint normal density \eqref{eq:discrete_normal_dist} by linear conditioning \cite[Theorem 1]{Pfortner.2022}.

\subsubsection{Continuous Lagrangian Gaussian Processes}

    Suitable normalization conditions are required to construct a continous-time \ac{lgp} from the \ac{gp} priors \eqref{eq:clgp_priors} and the linear operators ${\boldsymbol{\mathcal{L}}}_{\mathrm{DI}}$ and ${\boldsymbol{\mathcal{F}}}_{\mathrm{DI}}$ in \eqref{eq:clgp_operators}. 
    
	\paragraph{Normalization.} 
	For continuous Lagrangians to be non-degenerate (\ie regular), $\frac{\partial^2 L}{\partial \dot{\boldsymbol{q}} \partial \dot{\boldsymbol{q}}}$ needs to be invertible everywhere \cite{OberBlobaum.2011,Offen.2025}. 
	
	Similar to Section\,\ref{sec:dt_lgp}, we follow the lines of \cite{Offen.2025} and consider $2 n_q +1$ normalization conditions. 
	To this end, we enforce---at some anchor points $\bar{\boldsymbol{z}}_L$ and $\bar{\boldsymbol{x}}_F$---a fixed value $n_L \neq 0$ of the continuous Lagrangian through the evaluation operator $\mathcal{E}_{\bar{\boldsymbol{z}}_L} \left[L \right]  \triangleq L (\bar{\boldsymbol{z}}_L)$ and a fixed momentum $\boldsymbol{n}_M \neq \boldsymbol{0}$ through the momentum operator $\boldsymbol{\mathcal{M}}_{\bar{\boldsymbol{z}}_L} \left[L \right]  \triangleq \left. \frac{\partial L}{\partial \dot{\boldsymbol{q}} } \right|_{\bar{\boldsymbol{z}}_L}$. 
    Moreover, the condition $\boldsymbol{F}(\boldsymbol{0},\boldsymbol{0},\boldsymbol{0}) \stackrel{}{=} \boldsymbol{0} =: \boldsymbol{n}_F$ is imposed via the evaluation operator $\boldsymbol{\mathcal{E}}_{\bar{\boldsymbol{x}}_F}\left[ \boldsymbol{F}_{\Delta}^{\pm} \right] \triangleq \boldsymbol{F}_{\Delta}^{\pm} (\bar{\boldsymbol{x}}_F)$, which ensures vanishing external forces at rest if $\boldsymbol{u} = \boldsymbol{0}$.  

    We formulate these normalization conditions as additional linear operators which are appended---for a training data set of size $N$---to $N$ evaluations of \eqref{eq:clgp_operators}, \ie
	\begin{align}\label{eq:clgp_augm_el_operators}
		\bar{\boldsymbol{\mathcal{L}}}_{\mathrm{DI}} &\triangleq \begin{bmatrix} {\boldsymbol{\mathcal{L}}}_{\mathrm{DI}} \\ \vdots \\ {\boldsymbol{\mathcal{L}}}_{\mathrm{DI}}  \\ {\boldsymbol{\mathcal{C}}}_{{L\mathrm{I}}} \\ {\boldsymbol{0}} 
		\end{bmatrix} \, , \: 
		\bar{\boldsymbol{\mathcal{F}}}_{\mathrm{DI}} \triangleq \begin{bmatrix}
			{\boldsymbol{\mathcal{F}}}_{\mathrm{DI}} \\ \vdots \\  {\boldsymbol{\mathcal{F}}}_{\mathrm{DI}} \\ {\boldsymbol{0}}  \\ {\boldsymbol{\mathcal{C}}}_{{F\mathrm{I}}}
		\end{bmatrix} \, , \quad \mathrm{with} \quad
		{\boldsymbol{\mathcal{C}}}_{{L\mathrm{I}}} \left[L \right] \triangleq \begin{bmatrix} \mathcal{E}_{\bar{\boldsymbol{z}}_L} \left[L \right]  \\ \boldsymbol{\mathcal{M}}_{\bar{\boldsymbol{z}}_L} \left[L \right]
		\end{bmatrix} \, , \:
		{\boldsymbol{\mathcal{C}}}_{{F\mathrm{I}}} \left[\boldsymbol{F} \right] \triangleq 
		\boldsymbol{\mathcal{E}}_{\bar{\boldsymbol{x}}_F} \left[\boldsymbol{F} \right] \, ,
	\end{align}
    with the same pseudo-measurement vector $\bar{\boldsymbol{y}}^\top = \left[\boldsymbol{y}^\top, \dots, \boldsymbol{y}^\top, {{n}}_L ,  {\boldsymbol{n}}_M^\top , {\boldsymbol{n}}_F^\top\right] \in \mathbb{R}^{(N + 2) n_q +1}$ as in the discrete case. 
	The resulting joint normal distribution with normalization conditions is
	\begin{align}
		&\begin{bmatrix}
			L  \\ \boldsymbol{F} \\ \bar{\boldsymbol{y}} 
		\end{bmatrix} \sim \mathcal{N} \left(\begin{bmatrix}
			{0} \\ \boldsymbol{0} \\ \boldsymbol{0}  
		\end{bmatrix}, \begin{bmatrix}
			k_L(\boldsymbol{z},\boldsymbol{z}^{\prime}) & \boldsymbol{0} &  k_L \bar{\boldsymbol{\mathcal{L}}}_{\mathrm{DI}}^{\prime}  \\
			\boldsymbol{0} & \boldsymbol{k}_{F} \left(\boldsymbol{x}, \boldsymbol{x}^{\prime}\right) & \boldsymbol{k}_F \bar{\boldsymbol{\mathcal{F}}}_{\mathrm{DI}}^{\prime}  \\
			\bar{\boldsymbol{\mathcal{L}}}_{\mathrm{DI}} k_L & \bar{\boldsymbol{\mathcal{F}}}_{\mathrm{DI}} \boldsymbol{k}_F & \bar{\boldsymbol{\Theta}}_{\mathrm{DI}}  \\
		\end{bmatrix}\right)\, ,  
    \end{align}
    which ensures non-degeneracy of the learned continuous Lagrangian \cite{Offen.2025}. 
    The matrix $\bar{\boldsymbol{\Theta}}_{\mathrm{DI}} \in \mathbb{R}^{[(N+2)n_q +1] \times [(N+2)n_q +1]}$ is defined as
\begin{align}
\bar{\boldsymbol{\Theta}}_{\mathrm{DI}} \triangleq
\left[
\begin{array}{cccc|cc}
    \boldsymbol{\Theta}_{\mathrm{DI}, 11}+ \boldsymbol{\Sigma} & \boldsymbol{\Theta}_{\mathrm{DI}, 12} & \dots & \boldsymbol{\Theta}_{\mathrm{DI}, 1N} & \boldsymbol{\mathcal{L}}_{\mathrm{DI},1} k_L \boldsymbol{\mathcal{C}}_{L\mathrm{I}}^{\prime} & \boldsymbol{\mathcal{F}}_{\mathrm{DI},1}\boldsymbol{k}_{F}\boldsymbol{\mathcal{C}}_{F\mathrm{I}}^{\prime}\\
    \boldsymbol{\Theta}_{\mathrm{DI}, 21} & \boldsymbol{\Theta}_{\mathrm{DI},22}+ \boldsymbol{\Sigma} & \dots & \boldsymbol{\Theta}_{\mathrm{DI}, 2N} & \boldsymbol{\mathcal{L}}_{\mathrm{DI},2} k_L \boldsymbol{\mathcal{C}}_{L\mathrm{I}}^{\prime} & \boldsymbol{\mathcal{F}}_{\mathrm{DI},2}\boldsymbol{k}_{F}\boldsymbol{\mathcal{C}}_{F\mathrm{I}}^{\prime}\\
    \vdots & \vdots & \ddots & \vdots & \vdots & \vdots \\
    \boldsymbol{\Theta}_{\mathrm{DI}, N1} & \boldsymbol{\Theta}_{\mathrm{DI}, N2} & \dots & \boldsymbol{\Theta}_{\mathrm{DI},NN}+ \boldsymbol{\Sigma} & \boldsymbol{\mathcal{L}}_{\mathrm{DI},N} k_L \boldsymbol{\mathcal{C}}_{L\mathrm{I}}^{\prime} & \boldsymbol{\mathcal{F}}_{\mathrm{DI},N}\boldsymbol{k}_{F}\boldsymbol{\mathcal{C}}_{F\mathrm{I}}^{\prime}\\
    \hline
    \boldsymbol{\mathcal{C}}_{L\mathrm{I}} k_L \boldsymbol{\mathcal{L}}_{\mathrm{DI},1}^{\prime} & \boldsymbol{\mathcal{C}}_{L\mathrm{I}} k_L \boldsymbol{\mathcal{L}}_{\mathrm{DI},2}^{\prime} & \dots & \boldsymbol{\mathcal{C}}_{L\mathrm{I}} k_L \boldsymbol{\mathcal{L}}_{\mathrm{DI},N}^{\prime} & \boldsymbol{\mathcal{C}}_{L\mathrm{I}} k_L \boldsymbol{\mathcal{C}}_{L\mathrm{I}}^{\prime} & \boldsymbol{0} \\
    \boldsymbol{\mathcal{C}}_{F\mathrm{I}}\boldsymbol{k}_F\boldsymbol{\mathcal{F}}_{\mathrm{DI},1}^{\prime}  & \boldsymbol{\mathcal{C}}_{F\mathrm{I}}\boldsymbol{k}_F\boldsymbol{\mathcal{F}}_{\mathrm{DI},2}^{\prime} & \dots & \boldsymbol{\mathcal{C}}_{F\mathrm{I}}\boldsymbol{k}_F\boldsymbol{\mathcal{F}}_{\mathrm{DI},N}^{\prime} & \boldsymbol{0} & \boldsymbol{\mathcal{C}}_{F\mathrm{I}}\boldsymbol{k}_F \boldsymbol{\mathcal{C}}_{F\mathrm{I}}^{\prime}
\end{array}
\right]\, ,
\label{eq:theta_di_matrix_expanded}
\end{align}
with $\boldsymbol{\Theta}_{\mathrm{DI},ij} \in \mathbb{R}^{n_q \times n_q}$ being the covariance of residual dynamics at two data points $i$ and $j$, \ie
\begin{equation}
\boldsymbol{\Theta}_{\mathrm{DI},ij} \triangleq  \boldsymbol{\mathcal{F}}_{\mathrm{DI},i} \boldsymbol{k}_F \boldsymbol{\mathcal{F}}_{\mathrm{DI},j}^{\prime}  +  \boldsymbol{\mathcal{L}}_{\mathrm{DI},i} k_L \boldsymbol{\mathcal{L}}_{\mathrm{DI},j}^{\prime}\, , \qquad \forall i, j \in [1,N] \,.
\end{equation}
Thus, we require $N$ data triplets $\{\boldsymbol{q}_{n-1}^{(i)}, \boldsymbol{q}_{n}^{(i)}, \boldsymbol{q}_{n+1}^{(i)}\}_{i=1}^{N}$ with corresponding inputs $\{ \boldsymbol{u}_{n-1}^{(i)},\boldsymbol{u}_{n}^{(i)} \}_{i=1}^{N} $ to construct $\bar{\boldsymbol{\Theta}}_{\mathrm{DI}}$ in \eqref{eq:theta_di_matrix_expanded}.

	\paragraph{Posteriors.}
	The continuous-time \ac{gp} priors \eqref{eq:clgp_priors} can be conditioned on the discrete forced Euler-Lagrange equations using the linear operators $\bar{\boldsymbol{\mathcal{L}}}_{\mathrm{DI}}$ and $\bar{\boldsymbol{\mathcal{F}}}_{\mathrm{DI}}$ \cite{Pfortner.2022}. 
	The marginal posteriors are 
	\begin{equation}
		\begin{aligned}
			& \left\{ L  \mid  \bar{\boldsymbol{\mathcal{L}}}_{\mathrm{DI}}\left[L \right] + \bar{\boldsymbol{\mathcal{F}}}_{\mathrm{DI}}\left[\boldsymbol{F}\right] + \bar{\boldsymbol{\epsilon}} = \bar{\boldsymbol{y}} \right\}  \sim \mathcal{GP}\left( {m}^{L  | \bar{\boldsymbol{y}}} , {k}^{L  | \bar{\boldsymbol{y}}} \right) \, , \\
			&\hspace{1cm} {m}^{L  | \bar{\boldsymbol{y}}} (\boldsymbol{z})= \bar{\boldsymbol{\mathcal{L}}}_{\mathrm{DI}}\left[k_L\left(\boldsymbol{z},\cdot\right)\right]^{\top} \bar{\boldsymbol{\Theta}}_{\mathrm{DI}}^{\dagger} \bar{\boldsymbol{y}} \, ,\\
			&\hspace{1cm} {k}^{L  | \bar{\boldsymbol{y}}} (\boldsymbol{z}_1,\boldsymbol{z}_2) = k_L\left( \boldsymbol{z}_1,\boldsymbol{z}_2 \right)-  \bar{\boldsymbol{\mathcal{L}}}_{\mathrm{DI}}\left[k_L\left(\boldsymbol{z}_1,\cdot\right)\right]^{\top} \bar{\boldsymbol{\Theta}}_{\mathrm{DI}}^{\dagger} \bar{\boldsymbol{\mathcal{L}}}_{\mathrm{DI}}\left[k_L\left(\cdot,\boldsymbol{z}_2\right)\right]\, ,
		\end{aligned}
	\end{equation}
	for the Lagrangian, and the marginal posterior of the learned force is
	\begin{equation}
		\begin{aligned}
			& \left\{ \boldsymbol{F}  \mid  \bar{\boldsymbol{\mathcal{L}}}_{\mathrm{DI}}\left[L \right] + \bar{\boldsymbol{\mathcal{F}}}_{\mathrm{DI}}\left[\boldsymbol{F}\right] + \bar{\boldsymbol{\epsilon}} = \bar{\boldsymbol{y}} \right\}  \sim \mathcal{GP}\left( \boldsymbol{m}^{\boldsymbol{F}  | \bar{\boldsymbol{y}}} , \boldsymbol{k}^{\boldsymbol{F}  | \bar{\boldsymbol{y}}} \right) \, , \\
			&\hspace{1cm} \boldsymbol{m}^{\boldsymbol{F} | \bar{\boldsymbol{y}}} (\boldsymbol{x})= \bar{\boldsymbol{\mathcal{F}}}_{\mathrm{DI}}\left[\boldsymbol{k}_F\left(\boldsymbol{x},\cdot\right)\right]^{\top} \bar{\boldsymbol{\Theta}}_{\mathrm{DI}}^{\dagger} \bar{\boldsymbol{y}} \, ,\\
			&\hspace{1cm} \boldsymbol{k}^{\boldsymbol{F} | \bar{\boldsymbol{y}}} (\boldsymbol{x}_1,\boldsymbol{x}_2)= \boldsymbol{k}_F \left( \boldsymbol{x}_1,\boldsymbol{x}_2 \right)-  \bar{\boldsymbol{\mathcal{F}}}_{\mathrm{DI}}\left[\boldsymbol{k}_F\left(\boldsymbol{x}_1,\cdot\right)\right]^{\top} \bar{\boldsymbol{\Theta}}_{\mathrm{DI}}^{\dagger} \bar{\boldsymbol{\mathcal{F}}}_{\mathrm{DI}}\left[\boldsymbol{k}_F\left(\cdot,\boldsymbol{x}_2\right)\right]\, .
		\end{aligned}
	\end{equation}

\subsection{Prediction of Linear Observables}\label{app:linear_observables}

 Let $\boldsymbol{\mathcal{O}}$ be an arbitrary linear operator acting on the Lagrangian $L $. Since linear transformations of Gaussian processes remain Gaussian, the posterior distribution of the observable $\boldsymbol{\mathcal{O}}[L ]$ is given by
 \begin{align}
 \boldsymbol{\mathcal{O}}\left[L \right] \mid \bar{\boldsymbol{y}} &\sim \mathcal{GP}\left( m^{\boldsymbol{\mathcal{O}}[L] | \bar{\boldsymbol{y}}}, k^{\boldsymbol{\mathcal{O}}[L] | \bar{\boldsymbol{y}}} \right) \, .
 \end{align}
 The posterior mean function of the observable is obtained by applying the operator to the posterior mean of the Lagrangian
 \begin{align}
 m^{\boldsymbol{\mathcal{O}}[L] | \bar{\boldsymbol{y}}}(\boldsymbol{z}) &= \boldsymbol{\mathcal{O}} \left[ m^{L  | \bar{\boldsymbol{y}}} \right](\boldsymbol{z}) \nonumber \\
 &= \bar{\boldsymbol{\mathcal{L}}}_{\mathrm{DI}}\left[\boldsymbol{\mathcal{O}}_{\boldsymbol{z}}\left[k_L\left(\cdot, \boldsymbol{z}\right)\right]\right]^{\top} \bar{\boldsymbol{\Theta}}_{\mathrm{DI}}^{\dagger} \bar{\boldsymbol{y}} \, .
 \end{align}
 The posterior covariance function is derived by applying the operator to both arguments of the posterior kernel
 \begin{align}
 k^{\boldsymbol{\mathcal{O}}[L] | \bar{\boldsymbol{y}}}(\boldsymbol{z}_1, \boldsymbol{z}_2) &= \boldsymbol{\mathcal{O}}_{\boldsymbol{z}_1} \left[ \boldsymbol{\mathcal{O}}_{\boldsymbol{z}_2} \left[ k^{L  | \bar{\boldsymbol{y}}}(\boldsymbol{z}_1, \boldsymbol{z}_2) \right] \right] \nonumber \\
 &= \boldsymbol{\mathcal{O}}_{\boldsymbol{z}_1} \left[ \boldsymbol{\mathcal{O}}_{\boldsymbol{z}_2} \left[ k_L(\boldsymbol{z}_1, \boldsymbol{z}_2) \right] \right] - \boldsymbol{v}(\boldsymbol{z}_1)^\top \bar{\boldsymbol{\Theta}}_{\mathrm{DI}}^{\dagger} \boldsymbol{v}(\boldsymbol{z}_2) \, ,
 \end{align}
 where $\boldsymbol{v}(\boldsymbol{z}) = \bar{\boldsymbol{\mathcal{L}}}_{\mathrm{DI}}\left[ \boldsymbol{\mathcal{O}}\left[ k_L(\cdot, \boldsymbol{z}) \right] \right]$.
 Here, the subscript in $\boldsymbol{\mathcal{O}}_{\boldsymbol{z}_i}$ indicates that the linear operator acts on the argument $\boldsymbol{z}_i$ of the kernel, \ie $\boldsymbol{\mathcal{O}}_{\boldsymbol{z}_1}\left[k_L(\boldsymbol{z}_1,\boldsymbol{z}_2)\right] := \boldsymbol{\mathcal{O}}\left[\boldsymbol{z}\mapsto k_L(\boldsymbol{z},\boldsymbol{z}_2)\right](\boldsymbol{z}_1)$, while fixing the other argument.

 \paragraph{Application to the Hamiltonian.}
 The Hamiltonian for a Lagrangian system is defined by the Legendre transform $H(\boldsymbol{q}, \dot{\boldsymbol{q}}) = \dot{\boldsymbol{q}}^\top \frac{\partial L}{\partial \dot{\boldsymbol{q}}} - L$. This can be viewed as a linear operator $\boldsymbol{\mathcal{H}}$ acting on $L$, \ie
 \begin{equation}
 \boldsymbol{\mathcal{H}}[L](\boldsymbol{z}) = \dot{\boldsymbol{q}}^\top \nabla_{\dot{\boldsymbol{q}}} L(\boldsymbol{z}) - L(\boldsymbol{z}) \, .
 \end{equation}
 To compute the learned Hamiltonian from the \acp{lgp}, we apply the formulas above with $\boldsymbol{\mathcal{O}} := \boldsymbol{\mathcal{H}}$. 
 The predicted Hamiltonian mean is
 \begin{equation}
 \mu_{H}(\boldsymbol{z}) = \dot{\boldsymbol{q}}^\top \nabla_{\dot{\boldsymbol{q}}} m^{L  | \bar{\boldsymbol{y}}}(\boldsymbol{z}) - m^{L  | \bar{\boldsymbol{y}}}(\boldsymbol{z}) \, .
 \end{equation}
 The predicted variance of the Hamiltonian at a state $\boldsymbol{z}$ is
 \begin{align}
 \sigma^2_{H}(\boldsymbol{z}) &= \boldsymbol{\mathcal{H}}_{\boldsymbol{z}'} \left[ \boldsymbol{\mathcal{H}}_{\boldsymbol{z}} \left[ k_L(\boldsymbol{z}, \boldsymbol{z}') \right] \right]_{\boldsymbol{z}'=\boldsymbol{z}}  - \bar{\boldsymbol{\mathcal{L}}}_{\mathrm{DI}}\left[ \boldsymbol{\mathcal{H}}_{\boldsymbol{z}}\left[ k_L(\cdot, \boldsymbol{z}) \right] \right]^{\top} \bar{\boldsymbol{\Theta}}_{\mathrm{DI}}^{\dagger} \bar{\boldsymbol{\mathcal{L}}}_{\mathrm{DI}}\left[ \boldsymbol{\mathcal{H}}_{\boldsymbol{z}}\left[ k_L(\cdot, \boldsymbol{z}) \right] \right] \, .
 \end{align}
 Here, the term $\boldsymbol{\mathcal{H}}_{\boldsymbol{z}} \left[ k_L(\boldsymbol{z}', \boldsymbol{z}) \right]$ involves differentiating the kernel with respect to the velocity components of its second argument
 \begin{equation}
 \boldsymbol{\mathcal{H}}_{\boldsymbol{z}} \left[ k_L(\boldsymbol{z}', \boldsymbol{z}) \right] = \dot{\boldsymbol{q}}^\top \nabla_{\dot{\boldsymbol{q}}} k_L(\boldsymbol{z}', \boldsymbol{z}) - k_L(\boldsymbol{z}', \boldsymbol{z}) \, .
 \end{equation}

Therefore, Hamiltonian mean and covariance are obtained directly from the learned Lagrangian \ac{gp} posterior through linear operators, consistent with \cite{Offen.2025}.

	\subsection{Incorporating Energy Models in the Kernel}\label{sec:phys_kernel_structure}
	
	While the \acp{lgp} presented in Sections\,\ref{sec:dt_lgp} and \ref{sec:ct_lgp} \textit{are not restricted} to specific kernel designs, further knowledge about the energy model \textit{can} be incorporated to facilitate learning. 
	For instance, the Lagrangian of various mechanical rigid-body systems is composed of the kinetic energy $T: TQ \rightarrow \mathbb{R}$ and the potential energy $V: Q \rightarrow \mathbb{R}$ according to
	\begin{equation}\label{eq:lagrangian_knowledge}
		L(\boldsymbol{q},\dot{\boldsymbol{q}}) = \underbrace{\frac{1}{2} \dot{\boldsymbol{q}}^\top \boldsymbol{M}(\boldsymbol{q})\dot{\boldsymbol{q}}}_{T(\boldsymbol{q},\dot{\boldsymbol{q}})} - \underbrace{ \left( \frac{1}{2} \boldsymbol{q}^\top \boldsymbol{S}(\boldsymbol{q}){\boldsymbol{q}} + \boldsymbol{G}(\boldsymbol{q}) \right) }_{V(\boldsymbol{q})} \, ,
	\end{equation}
	where the mass matrix $\boldsymbol{M}: Q \rightarrow \mathbb{R}^{n_q \times n_q}$ and the stiffness matrix $\boldsymbol{S}: Q \rightarrow \mathbb{R}^{n_q \times n_q}$ are symmetric positive definite. 
	Moreover, the gravitational term $\boldsymbol{G} : Q \rightarrow \mathbb{R}$ has usually an equilibrium, \ie $\boldsymbol{G}(\boldsymbol{0})=0$ and $\nabla_{\boldsymbol{q}} \boldsymbol{G}(\boldsymbol{0})= \boldsymbol{0}$ \cite{Evangelisti.2022b}. 
	As for the external force $\boldsymbol{F}$, common modeling choices are a Raleigh dissipation function $D(\boldsymbol{q},\dot{\boldsymbol{q}}) = \frac{1}{2}\dot{\boldsymbol{q}}^\top  \boldsymbol{C}(\boldsymbol{q}) \dot{\boldsymbol{q}}$ \cite{Goldstein.2002} and affine inputs, \ie
	\begin{equation}\label{eq:force_knowledge}
		\boldsymbol{F}(\boldsymbol{u},\boldsymbol{q},\dot{\boldsymbol{q}}) =  \boldsymbol{B}(\boldsymbol{q}) {\boldsymbol{u}} - \underbrace{\boldsymbol{C}(\boldsymbol{q}) \dot{\boldsymbol{q}}}_{\nabla_{\dot{\boldsymbol{q}}} \boldsymbol{D}(\boldsymbol{q},\dot{\boldsymbol{q}}) } \, ,
	\end{equation}
	where $\boldsymbol{C}: Q \rightarrow \mathbb{R}^{n_q \times n_q}$ and $\boldsymbol{B}: Q \rightarrow \mathbb{R}^{n_q \times n_q}$ are positive semi-definite matrices. 
	System knowledge, such as the linear and quadratic terms in \eqref{eq:lagrangian_knowledge} and \eqref{eq:force_knowledge}, can be incorporated in the \ac{gp} kernels using homogeneous polynomial covariance functions \cite{Rasmussen.2005}. 
	For example, suitable kernel designs for the dissipation term and the kinetic energy would be
		$k_D(\boldsymbol{z},\boldsymbol{z}^{\prime}) = k(\boldsymbol{q},\boldsymbol{q}^{\prime}) \cdot  \dot{\boldsymbol{q}}^\top \dot{\boldsymbol{q}}^{\prime}$ and
		$k_T(\boldsymbol{z},\boldsymbol{z}^{\prime}) = k(\boldsymbol{q},\boldsymbol{q}^{\prime}) \cdot \left( \dot{\boldsymbol{q}}^\top \dot{\boldsymbol{q}}^{\prime}\right)^2 $, 
	where $k(\boldsymbol{q},\boldsymbol{q}^{\prime})$ is again a generic kernel that models the contributions of the matrices $\boldsymbol{M}(\boldsymbol{q})$ and $\boldsymbol{C}(\boldsymbol{q})$, respectively. 
	Summing the respective kernels yields the overall covariance functions for use in \eqref{eq:dlgp_priors} and \eqref{eq:clgp_priors}. 
	We stress that this procedure for incorporating prior knowledge does not ensure positive definite matrices $\boldsymbol{M}$ and $\boldsymbol{S}$.

\subsection{Kernel Design Guide: Generalization and Ambiguity of Lagrangians}\label{app:ambiguity_of_Lagrangians}

In this appendix, we illustrate how different levels of prior knowledge in the kernel design affect the learning and prediction performance. 
Specifically, we demonstrate that the obtained Lagrangian can be ambiguous, despite having learned a unique Euler-Lagrange operator, which is useful for prediction tasks. 
Moreover, we demonstrate that adding prior knowledge to the kernel design improves generalization. 
To this end, we compare three continuous \acp{lgp} with different kernels, one with a \emph{generic} squared exponential kernel, one with a \emph{linear force kernel} according to \eqref{eq:force_knowledge}, and one with a full \emph{physics-inspired} kernel (see Appendix\,\ref{sec:phys_kernel_structure}), in the controlled single pendulum simulation from Figure\,\ref{fig:pendulum_structure_Lagrangians} in Section\,\ref{sec:results_sim_pendulum}. 
Further details about the simulation and the kernel choice are given in Appendix\,\ref{app:multi_link_pendulum}.

\begin{figure}[tb]
	\centering
    \includegraphics[width=0.99\textwidth]{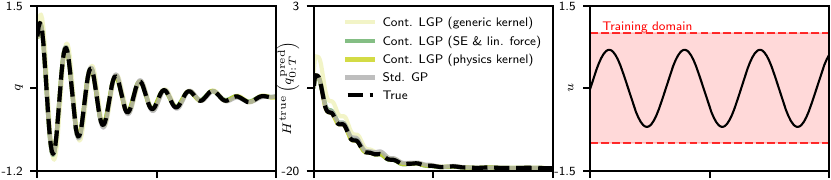}
	\includegraphics[width=0.99\textwidth]{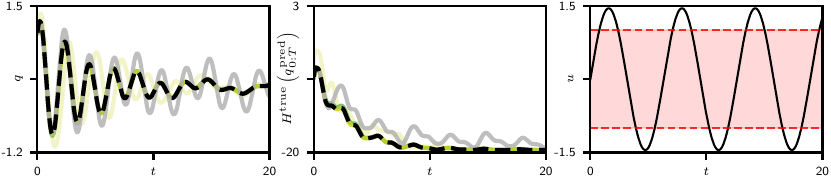}
	\caption{Predicted trajectories of continuous \acp{lgp} with different levels of prior knowledge in the kernel design. The figure shows trajectories of a controlled pendulum (left) and corresponding system energy (Hamiltonian $H$), evaluated at the predictions (middle). The control input (right) is chosen either \emph{in-distribution} (top) or \emph{out-of-distribution} (bottom).}
	\label{fig:app_ambiguity_sim_pendulum_trajectories}
\end{figure}

\paragraph{Forward predictions.} 
First, we are interested in performing multi-step forward predictions, given two initial angle coordinates $q_0$ and $q_1$ as well as control input $u$. 
In Figure\,\ref{fig:app_ambiguity_sim_pendulum_trajectories}, we evaluate two scenarios, one \emph{in-distribution} case in which the input $u$ stays within the region of the training data and one \emph{out-of-distribution} scenario in which $u$ exceeds the training domain. 
As shown in the figure, the \acp{lgp} provide accurate long-term forward predictions \emph{in-distribution}, despite varying levels of prior knowledge in the kernel design. 
Evaluated \emph{out-of-distribution}, we see that the \acp{lgp} with physics kernels still provide highly accurate predictions, while the predictive performance of the \ac{lgp} with a generic kernel and a baseline \ac{gp} degrades. 
This result suggests that the proposed \acp{lgp} yield highly accurate multi-step predictions with both generic and physics-inspired kernels. 
Still, adding prior knowledge about a system's energy model can improve the predictive performance, especially \emph{out-of-distribution}. 

\begin{figure}[tb]
	\centering
    \begin{subfigure}{0.32\textwidth}
        \centering
        \includegraphics[height=6cm]{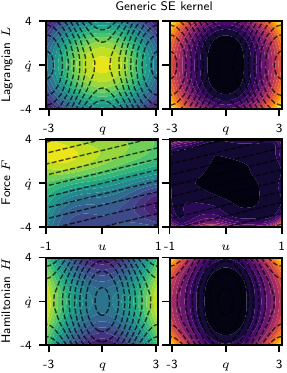}
        \caption{\ac{lgp} with generic squared exponential kernel for both $L$ and $F$.}
        \label{fig:app_ambiguity_sim_pendulum_oscillator1}
    \end{subfigure}%
    \hfill
    \begin{subfigure}{0.32\textwidth}
        \centering
        \includegraphics[height=6cm]{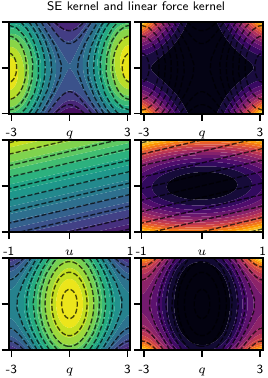}
        \caption{\ac{lgp} with squared exp. kernel for $L$ and linear kernel acc. to \eqref{eq:force_knowledge} for $F$.}
        \label{fig:app_ambiguity_sim_pendulum_oscillator2}
    \end{subfigure}%
    \hfill
    \begin{subfigure}{0.32\textwidth}
        \centering
        \includegraphics[height=6cm]{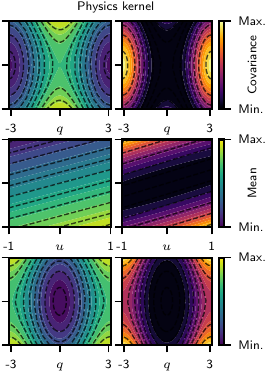}
        \caption{\ac{lgp} with full physics-inspired kernel acc. to \eqref{eq:lagrangian_knowledge} and \eqref{eq:force_knowledge}.}
        \label{fig:app_ambiguity_sim_pendulum_oscillator3}
    \end{subfigure}%
	\caption{Learned Lagrangian, external force, and Hamiltonian mean and covariance in a single pendulum example. The subplots compare continuous \acp{lgp} with different levels of prior knowledge in the kernel design ($N=300$, $n_q=1$, dashed: true contours).}
	\label{fig:app_ambiguity_sim_pendulum_oscillator}
\end{figure}

\paragraph{Learned dynamics model.} 
{Second}, we are interested in learning the external force $F$, the Lagrangian $L$, and the Hamiltonian $H$ of the pendulum. 
Figure\,\ref{fig:app_ambiguity_sim_pendulum_oscillator} summarizes the learned mean and covariance functions for both \acp{lgp}. 
As expected, the \ac{lgp} with physics-inspired kernel in Figure\,\ref{fig:app_ambiguity_sim_pendulum_oscillator3} matches the shape of the true quantities, while the variant with a generic kernel in Figure\,\ref{fig:app_ambiguity_sim_pendulum_oscillator1} does not. 
However, inspecting the learned quantities more closely, we see that the learned Lagrangian $L$ in the generic-kernel case qualitatively reflects the true Hamiltonian. 
Vice versa, the corresponding learned Hamiltonian $H$ qualitatively reflects the true Lagrangian. 
This observation is insightful, as it hints at Ambiguity in the learned Lagrangians. 

\paragraph{Ambiguity in unforced Euler-Lagrange equations.} 
The reason for this ambiguity is that the normalization conditions in Section\,\ref{sec:methods} rule out degenerate Lagrangians, but do not restrict learning to the exact shape of the true Lagrangian. 
For \emph{unforced} systems, \cite[Section 2.1.2]{Offen.2025} characterizes the residual non-uniqueness via the equivalence class
\begin{align}\label{eq:equiv_Lagrangians}
    \tilde{L} (\boldsymbol{q}, \dot{\boldsymbol{q}}) &= \rho \, {L}(\boldsymbol{q}, \dot{\boldsymbol{q}}) + c + \dot{\boldsymbol{q}}^\top \nabla  W(\boldsymbol{q})\, ,
\end{align}
parameterized by a continuously differentiable function $W: \mathbb{R}^{n_q} \rightarrow \mathbb{R}$ and constants $\rho \in \mathbb{R} \setminus \{0\}$, $c \in \mathbb{R}$, provided that there does not exist an \emph{alternative} Lagrangian that explains the data. 
For the Hamiltonian, \eqref{eq:equiv_Lagrangians} induces
\begin{align}\label{eq:equiv_Hamiltonians}
    \boldsymbol{\mathcal{H}}[\tilde{L}](\boldsymbol{q}, \dot{\boldsymbol{q}}) &= \rho \, \boldsymbol{\mathcal{H}}[{L}](\boldsymbol{q}, \dot{\boldsymbol{q}}) - c \, , 
\end{align}
according to \cite[Lemma 2.3]{Offen.2025}.
Similar results for other linear observables and discrete Lagrangians are given in \cite{Offen.2025}.

\paragraph{Ambiguity in forced Euler-Lagrange equations.}
However, the equivalence class \eqref{eq:equiv_Lagrangians} alone does not capture all ambiguous solutions in our \emph{forced} setting.
In particular, we face two challenges: First, the data that is generated by a system that obeys the forced Euler-Lagrange equations
\begin{align}\label{eq:forced_EL}
    \frac{\mathrm{d}}{\mathrm{d}t} \nabla_{\dot{\boldsymbol{q}}} L(\boldsymbol{q}, \dot{\boldsymbol{q}}) - \nabla_{\boldsymbol{q}} L(\boldsymbol{q}, \dot{\boldsymbol{q}}) &= \boldsymbol{F}(\boldsymbol{u}, \boldsymbol{q}, \dot{\boldsymbol{q}}) \, ,
\end{align}
and second, that the Lagrangian and external force are learned \emph{jointly}.
In this case, for any continuously differentiable $\Phi: \mathbb{R}^{n_q} \rightarrow \mathbb{R}$, the modified pair $(\mathring{L}, \mathring{\boldsymbol{F}})$, that is 
\begin{align}\label{eq:LF_gauge}
    \mathring{L}(\boldsymbol{q}, \dot{\boldsymbol{q}}) &= L(\boldsymbol{q}, \dot{\boldsymbol{q}}) + \Phi(\boldsymbol{q})\, , 
    & \mathring{\boldsymbol{F}}(\boldsymbol{u}, \boldsymbol{q}, \dot{\boldsymbol{q}}) &= \boldsymbol{F}(\boldsymbol{u}, \boldsymbol{q}, \dot{\boldsymbol{q}}) - \nabla_{\boldsymbol{q}} \Phi(\boldsymbol{q}) \, ,
\end{align}
yields the same trajectories as $(L, \boldsymbol{F})$ under \eqref{eq:forced_EL}, since both sides of \eqref{eq:forced_EL} can be shifted by $-\nabla_{\boldsymbol{q}} \Phi$ while still satisfying the quality.

For mechanical Lagrangians $L = T - V$, the choice $\Phi(\boldsymbol{q}) = 2\,V(\boldsymbol{q})$ in \eqref{eq:LF_gauge} maps the true Lagrangian onto its energy,
\begin{align}\label{eq:Lt_HplusV}
    \mathring{L}(\boldsymbol{q}, \dot{\boldsymbol{q}}) &= T(\boldsymbol{q}, \dot{\boldsymbol{q}}) + V(\boldsymbol{q})\, , 
    & \mathring{\boldsymbol{F}}(\boldsymbol{x}) &= \boldsymbol{F}(\boldsymbol{x}) - 2 \nabla_{\boldsymbol{q}} V(\boldsymbol{q})\, ,
\end{align}
which numerically coincides with $\boldsymbol{\mathcal{H}}[L]$, while the corresponding Hamiltonian satisfies $\boldsymbol{\mathcal{H}}[\mathring{L}] = T - V = L$. 
In numerical studies, such as in Figure\,\ref{fig:app_ambiguity_sim_pendulum_oscillator1}, we observed this swap of learned Lagrangian and Hamiltonian multiple times when the force kernel was flexible enough to absorb $-2 \nabla_{\boldsymbol{q}} V$.  

To re-establish that the learned Lagrangians lie within the equivalence class \eqref{eq:equiv_Lagrangians}, \emph{additional} restrictions on the employed kernels are required. 
One example, which we use in this paper, is to prevent the force from absorbing terms of the form $\nabla_{\boldsymbol{q}} \Phi$ by employing a physics-informed kernel design, based on \eqref{eq:force_knowledge}. 
The effect of a linear force kernel can be seen in Figure\,\ref{fig:app_ambiguity_sim_pendulum_oscillator2}, where the learned quantities comply with the equivalence class \eqref{eq:equiv_Lagrangians} for some $\rho<0$. 
Adding further prior knowledge \eqref{eq:lagrangian_knowledge} about the Lagrangian rules out this sign ambiguity (see Figure\,\ref{fig:app_ambiguity_sim_pendulum_oscillator3}).

\newpage
\section{Experimental Details}\label{app:experimental_details}

This section provides extended methodological and experimental details.
We describe the training setup for each method (\cref{app:training_details}), the case studies used in our experiments (\cref{app:simulators}), and the evaluation metrics (\cref{app:metrics}).

\subsection{Training Setup} \label{app:training_details}

\paragraph{Lagrangian Gaussian processes.}
For both the discrete- and continuous \acp{lgp}, we use triplets of measured positions and corresponding inputs for training, \ie
\begin{equation}
	\left\{(\boldsymbol{q}_{n-1}^{(i)},\boldsymbol{q}_{n}^{(i)},\boldsymbol{q}_{n+1}^{(i)},\boldsymbol{u}_{n-1}^{(i)},\boldsymbol{u}_{n}^{(i)})\right\}_{i=1}^{N}.
\end{equation}
The target quantity is not the next position, but rather the discrete forced Euler-Lagrange equation \eqref{eq:DEL}. 
As the right-hand-side of \eqref{eq:DEL} is zero, the \acp{gp} are trained with the pseudo-measurements
\begin{equation}
	\bar{\boldsymbol{y}}^\top = \left[\boldsymbol{y}^\top, \dots, \boldsymbol{y}^\top, n_L,\; \boldsymbol{n}_M^\top,\; \boldsymbol{n}_F^\top\right],
\end{equation}
which is of dimension $(N + 2) \times n_q + 1$, as it contains $2 n_q +1$ normalization conditions to ensure regularity of the learned Lagrangians (cf. Sections~\ref{sec:dt_lgp} and \ref{sec:ct_lgp}).

The kernel hyperparameters \((\boldsymbol{\theta}_L,\boldsymbol{\theta}_F)\) of the Lagrangian and force \acp{gp} are set using \ac{ard} \cite{Rasmussen.2005,Neal.1996}. 
For hyperparameter optimization, we minimize negative log marginal likelihood with the L-BFGS-B optimizer and multiple restarts. 
The corresponding objective is
\begin{equation}
	\mathcal{J}(\log \boldsymbol{\theta}) 
	= \frac{1}{2}\bar{\boldsymbol{y}}^\top \bar{\boldsymbol{\Theta}}^{-1}\bar{\boldsymbol{y}}
	+ \frac{1}{2}\log\det(\bar{\boldsymbol{\Theta}})
	+ \mathcal{R}_{\mathrm{MAP}}(\log \boldsymbol{\theta}),
\end{equation}
where $\bar{\boldsymbol{\Theta}}$ is the augmented covariance matrix from \eqref{eq:theta_matrix_expanded} or \eqref{eq:theta_di_matrix_expanded}. $\mathcal{R}_{\mathrm{MAP}}$ is a Gaussian prior penalty on $\log \boldsymbol{\theta}$ to prioritize physically plausible values. 
We highlight that $\mathcal{R}_{\mathrm{MAP}}$ does not require system-specific prior knowledge. 

After selecting \((\boldsymbol{\theta}_L,\boldsymbol{\theta}_F)\), we perform a line search to fix the slack variable $\sigma$ in $\boldsymbol{\Sigma}:=\sigma \, \mathbf{I}$. 
To this end, we minimize the prediction error in a multi-step rollout on a short held-out trajectory. 
Finally, the model is re-fit once with \((\boldsymbol{\theta}_L^\star,\boldsymbol{\theta}_F^\star,\sigma^\star)\).
Please note that we standardize the kernel inputs component-wise using training statistics.

\paragraph{Standard Gaussian process.}
The baseline \ac{gp} is trained as a one-step predictor with the same local information as the \acp{lgp}. 
To this end, we construct the \ac{gp} features 
\begin{equation}
	\boldsymbol{p}_n = \left[\boldsymbol{q}_{n-1}^\top,\boldsymbol{q}_n^\top,\boldsymbol{u}_{n-1}^\top,\boldsymbol{u}_n^\top\right]^\top \, ,
\end{equation}
and the target $\boldsymbol{y}_n = \boldsymbol{q}_{n+1}$. 
Both features and targets are standardized with the statistics of the training data set. 
We use a separable kernel of the form
\begin{equation}
	k(\boldsymbol{p}_n,\boldsymbol{p}_n')=\sigma_k^2\exp\!\left(-\frac{1}{2}\sum_j\frac{(p_{n,j}-p_{n,j}')^2}{\ell_j^2}\right) \, ,
\end{equation}
where $p_{n,j}$ is the $j$-th entry of $\boldsymbol{p}_n$. 
Hyperparameters are obtained by maximizing the marginal likelihood with the L-BFGS-B optimizer and multiple restarts.

\newpage
\subsection{Case Studies} \label{app:simulators}

In this appendix, we detail the setup of the case studies in Section\,\ref{sec:results}.

\subsubsection{Multi-Link Pendulum Simulation} \label{app:multi_link_pendulum}

\begin{wrapfigure}{r}{0.35\textwidth} 
    \vspace{-5mm}
    \centering
    {\fontsize{6pt}{6pt}\selectfont
        \resizebox{0.35\textwidth}{!}{
            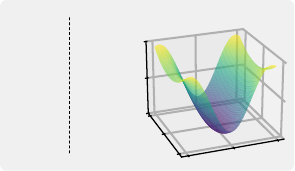
        }
    }
    \vspace{-6mm}
    \caption{Pendulum system and its energy function (Hamiltonian) for $n_q=1$.}
    \label{fig:app_pendulum}
    \vspace{-10mm}
\end{wrapfigure} 
\paragraph{System setup.} For the quantitative simulation study, we consider controlled, damped multi-link pendulums with the absolute angles with respect to the vertical axis as generalized coordinates $\boldsymbol{q}\in\mathbb{R}^{d}$, $n_q \in\{1,2,3\}$ (see Figure\,\ref{fig:app_pendulum}). 
The corresponding Lagrangian reads
\begin{equation}\label{eq:app_pendulum_lagrangian}
    \begin{aligned}
	L(\boldsymbol{q},\dot{\boldsymbol{q}})
	&=
	T(\boldsymbol{q},\dot{\boldsymbol{q}})-V(\boldsymbol{q}) \\
	&=
	\frac12\sum_{i=1}^{n_q} m_i\!\left(\left({v_{i}^{\mathrm{h}}}(\boldsymbol{q},\dot{\boldsymbol{q}})\right)^2+ \left({v_{i}^{\mathrm{v}}}(\boldsymbol{q},\dot{\boldsymbol{q}})\right)^2\right) \\
	& \quad -
	\sum_{i=1}^{n_q} m_i g\, h_i (\boldsymbol{q}) \, ,
    \end{aligned}
\end{equation}
where the vertical and horizontal velocities ${v_{i}^{\mathrm{v}}}$ and ${v_{i}^{\mathrm{h}}}$ of body $i$, as well as its height $h_i$ are determined by the kinematics
\begin{equation}\label{eq:app_pendulum_kinematics}
	{v_{i}^{\mathrm{h}}}(\boldsymbol{q},\dot{\boldsymbol{q}})=\sum_{j=1}^{i} l_j \dot q_j \cos q_j\, ,
	\qquad
	{v_{i}^{\mathrm{v}}}(\boldsymbol{q},\dot{\boldsymbol{q}})=\sum_{j=1}^{i} l_j \dot q_j \sin q_j\, ,
	\qquad
	h_i(\boldsymbol{q})=\sum_{j=1}^{i}(-l_j\cos q_j) \, .
\end{equation}
In \eqref{eq:app_pendulum_lagrangian} and \eqref{eq:app_pendulum_kinematics}, the masses and lengths are set linearly decreasing with the number of joints, \ie $m_i \in [2,\,1]$, $l_i \in [1,\,0.8]$.

The external forces in the pendulums are defined as the force differences at the respective joints $F_i = \tau_i^{\text{net}}-\tau_{i+1}^{\text{net}}$, $i=1,\dots,{n_q}$ with  
\begin{equation}\label{eq:app_pendulum_forces}
	\tau_i^{\text{net}} = u_i - b\,\omega_i \, , \quad \text{and} \quad \omega_i = \dot q_i-\dot q_{i-1} \, ,
\end{equation}
where $b\,\omega_i$ are damping terms with $b:=0.8$ that are linearly dependent on the relative angle velocities, and $\boldsymbol{u}^\top = [u_1, \dots, u_{n_q} ]$ are torque inputs at the respective joints. 
The boundary condition in \eqref{eq:app_pendulum_forces} is $\tau_{{n_q}+1}^{\text{net}}=0$. 
The ground-truth trajectories are generated by solving the discrete forced Euler-Lagrange equations \eqref{eq:DEL} using the midpoint rule and a root-finding algorithm.

\paragraph{Training data.}
To generate training data, we draw $N \in \{25,50,100,200,300\}$ random triplets
\begin{equation}
\left\{\boldsymbol{q}_{n-1}^{(i)},\boldsymbol{q}_{n}^{(i)},\boldsymbol{q}_{n+1}^{(i)}\right\}_{i=1}^{N},
\end{equation}
together with corresponding input samples, over the hypercube
$q_j\in[-\pi,\pi]$, $\dot q_j\in[-4,4]$, and $u_j\in[-1,1]$, for each dimension $j$. 
For the quantitative study in Figure\,\ref{fig:pendulum_structure_prediction}, we used a sampling time of $\Delta t_{\mathrm{train}} = 5 \cdot 10^{-2}$.

\paragraph{Normalization and kernel design.}
We consider two types of kernels, a generic force and Lagrangian kernels, as well as physics-inspired kernels (see Appendix\,\ref{sec:phys_kernel_structure}). 
The \textit{physics-inspired} Lagrangian kernel mirrors the energy decomposition \eqref{eq:lagrangian_knowledge} as
\begin{equation}\label{eq:app_pendulum_kernel_L}
	k_L(\boldsymbol{z},\boldsymbol{z}^{\prime}) = k_M(\boldsymbol{q},\boldsymbol{q}^{\prime})\,\left(\dot{\boldsymbol{q}}^{\top}\dot{\boldsymbol{q}}^{\prime}\right)^{2} + k_G(\boldsymbol{q},\boldsymbol{q}^{\prime}) + k_S(\boldsymbol{q},\boldsymbol{q}^{\prime})\,\left(\boldsymbol{q}^{\top}\boldsymbol{q}^{\prime}\right)^{2}\, ,
\end{equation}
with anisotropic squared exponential factors $k_M, k_G, k_S$ acting on $\boldsymbol{q}$ that model the mass-, gravitational-, and stiffness-related contributions to $L$, respectively.
The corresponding \textit{physics-inspired} force kernel encodes the control- and velocity-affine forces in \eqref{eq:force_knowledge} as
\begin{equation}\label{eq:app_pendulum_kernel_F}
	\bar{k}_F(\boldsymbol{x},\boldsymbol{x}^{\prime}) = k_R(\boldsymbol{q},\boldsymbol{q}^{\prime})\,\left(\boldsymbol{u}^{\top}\boldsymbol{u}^{\prime}\right) + k_D(\boldsymbol{q},\boldsymbol{q}^{\prime})\,\left(\dot{\boldsymbol{q}}^{\top}\dot{\boldsymbol{q}}^{\prime}\right)\, ,
\end{equation}
with the anisotropic squared exponential kernels $k_R$ and $k_D$ depending only on $\boldsymbol{q}$.
The \acp{lgp} with \textit{generic} kernels drop the polynomial factors and use a single anisotropic squared exponential kernel on the full Lagrangian state $\boldsymbol{z}=[\boldsymbol{q}^{\top},\dot{\boldsymbol{q}}^{\top}]^{\top}$ and on the full force input $\boldsymbol{x}=[\boldsymbol{u}^{\top},\boldsymbol{q}^{\top},\dot{\boldsymbol{q}}^{\top}]^{\top}$. 
For discrete kernels, $\dot{\boldsymbol{q}}$ is represented by $(\boldsymbol{q}_{n+1}-\boldsymbol{q}_n)/\Delta t_{\mathrm{train}}$, which aligns with the employed midpoint rule in Section\,\ref{sec:ct_lgp}.

Following Sections\,\ref{sec:dt_lgp} and \ref{sec:ct_lgp}, we enforce the pseudo-measurements $\bar{\boldsymbol{y}}^{\top}=\left[\boldsymbol{y}^{\top},\dots, \boldsymbol{y}^{\top},\,n_L,\,\boldsymbol{n}_M^{\top},\,\boldsymbol{n}_F^{\top}\right]$ as normalization conditions, with $n_L=1$ and $\boldsymbol{n}_F=\boldsymbol{0}$ throughout. 
For the \acp{lgp} with \textit{physics-inspired} kernels, the anchor is placed at the origin, \ie $\bar{\boldsymbol{q}}_{n-1}=\bar{\boldsymbol{q}}_{n}=\bar{\boldsymbol{q}}_{n+1}=\boldsymbol{0}$ with $\boldsymbol{n}_M=\boldsymbol{0}$, as the polynomial kinetic-energy term in the kernel itself already prevents null-Lagrangian solutions.
For the \acp{lgp} with \textit{generic} kernels, we additionally perturb the anchor by $\bar{\boldsymbol{q}}_{n\pm 1}=\pm 10^{-2}\,\boldsymbol{\sigma}_{\dot{\boldsymbol{q}}}\,\Delta t_{\mathrm{train}}$ and impose a non-zero momentum $\boldsymbol{n}_M=10^{-2}\,\boldsymbol{\sigma}_{\dot{\boldsymbol{q}}}$ to rule out null-Lagrangian solutions, where $\boldsymbol{\sigma}_{\dot{\boldsymbol{q}}}$ collects the per-component standard deviations of the finite-difference velocities in the training data. 

\paragraph{Test data.}
For each evaluation, we generate $50$ random forward-prediction scenarios, each with a $ 20$-step simulation horizon.
Each rollout is initialized from random $(\boldsymbol{q}_0,\dot{\boldsymbol{q}}_0)$, converted to two initial positions $(\boldsymbol{q}_{0},\boldsymbol{q}_{1})$, and driven by sinusoidal torques with random phase, frequency, and amplitude.

\subsubsection{Real-World Double Pendulum} \label{app:rw_pendulum}

\paragraph{System setup.} 
To test with real-world data, we consider the double-pendulum platform presented in \cite{Wiebe.2024} and shown in Figure\,\ref{fig:real_world_pendulum}. 
In particular, we evaluate on the system-identification records from ``\href{https://github.com/dfki-ric-underactuated-lab/double_pendulum/tree/main/data/experiment_records/design_C.1/sys_id/20230629}{design~C.1}'' in \cite{Kumar.2025} (the GitHub repository of \cite{Wiebe.2024}). 
The generalized coordinates of the double pendulum are the two joint angles $\boldsymbol{q}=[ q_1, q_2 ]^{\top}$. 
The double pendulum is driven by the motor torques $\boldsymbol{u}=[\tau_1, \tau_2]^{\top}$.
The system is non-conservative due to friction effects and actuator inputs. 
Moreover, it exhibits strongly nonlinear coupled dynamics, as is characteristic of underactuated link systems. 
We note that we do not assume any system-specific knowledge of the double-pendulum dynamics. 

\begin{figure}[tb]
	\centering
	\includegraphics[width=1\textwidth]{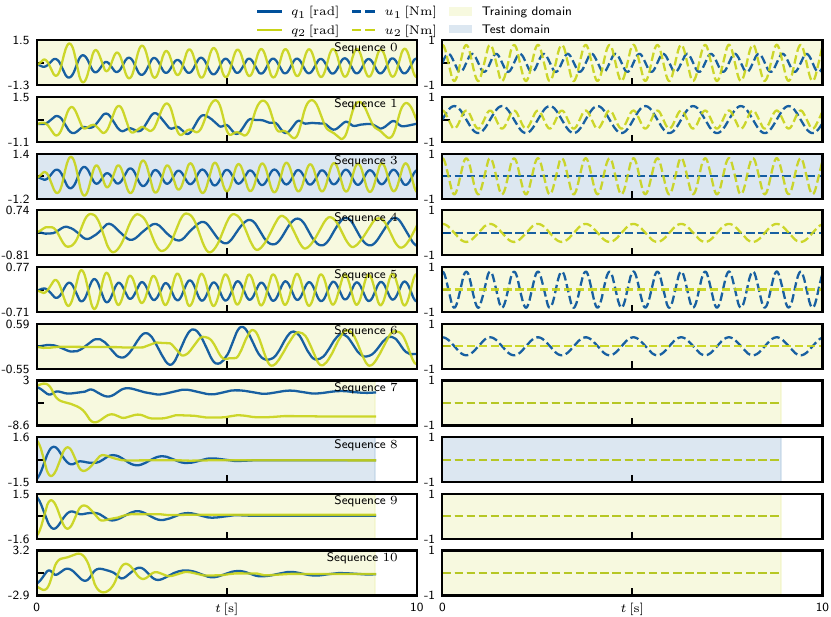}
	\caption{Training and test data sequences in the real-world double pendulum case study. Data set available from \cite{Wiebe.2024,Kumar.2025}.}
	\label{fig:app_rw_pendulum_sequences}
\end{figure}

\begin{figure}[tb]
	\centering
	\includegraphics[width=0.7\textwidth]{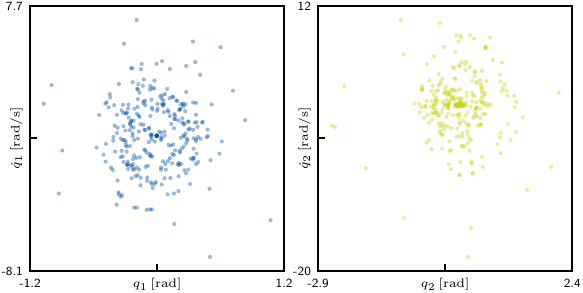}
	\caption{Scatter plot of the randomly drawn training data triplets. Each point in the plot represents a triplet.}
	\label{fig:app_rw_pendulum_scatter}
\end{figure}

\paragraph{Training data.} 
To generate the training data set, we sub-sample the recorded trajectories by a factor of $30$, which results in an effective training step size of $\Delta t_{\mathrm{train}} \approx 61~\mathrm{ms}$. 
We then randomly draw $N=300$ random triplets 
\begin{equation}
	\left\{\boldsymbol{q}_{n-1}^{(i)},\boldsymbol{q}_{n}^{(i)},\boldsymbol{q}_{n+1}^{(i)}\right\}_{i=1}^{N} \, ,
\end{equation}
together with the corresponding inputs. 
A single training data segment of $100$ steps is held out for tuning the scalar slack parameter $\sigma$ in $\boldsymbol{\Sigma} := \sigma \, \mathbf{I}$ and \eqref{eq:theta_matrix_expanded} by a line search. 
The resulting training data distribution is visualized in Figures\,\ref{fig:app_rw_pendulum_sequences} and \ref{fig:app_rw_pendulum_scatter}.

\paragraph{Normalization and kernel design.}
We train both the continuous and discrete \acp{lgp} using a \textit{physics-inspired} kernel design.  
The physics-inspired Lagrangian kernel follows the decomposition in Appendix~\ref{sec:phys_kernel_structure}, that is
\begin{equation}
	k_L = k_M(\boldsymbol{q},\boldsymbol{q}')\,(\dot{\boldsymbol{q}}^{\top}\dot{\boldsymbol{q}}')^2
	+ k_G(\boldsymbol{q},\boldsymbol{q}')
	+ k_S(\boldsymbol{q},\boldsymbol{q}')\,(\boldsymbol{q}^{\top}\boldsymbol{q}')^2 \, ,
\end{equation}
with anisotropic squared exponential sub-kernels $k_M,k_G,k_S$.  
The force kernel is
\begin{equation}
	\bar{k}_F = k_R(\boldsymbol{q},\boldsymbol{q}')\,(\boldsymbol{u}^{\top}\boldsymbol{u}')
	+ k_D(\boldsymbol{q},\boldsymbol{q}')\,(\dot{\boldsymbol{q}}^{\top}\dot{\boldsymbol{q}}') \, ,
\end{equation}
where $k_R$ and $k_D$ are squared exponential kernels. 
In the discrete \ac{lgp}, velocity factors are approximated as finite differences, $\dot{\boldsymbol{q}}\approx(\boldsymbol{q}_{n+1}-\boldsymbol{q}_n)/\Delta t_{\mathrm{train}}$, which aligns with the employed midpoint rule in Section\,\ref{sec:ct_lgp}. 

The normalization conditions are imposed as described in Section\,\ref{sec:dt_lgp}. 
For the Lagrangian, we use the anchor point $\bar{\boldsymbol{z}}_L = \bar{\boldsymbol{r}}_L = (\mathbf{0}, \mathbf{0})$ with targets $\boldsymbol{n}_M= \boldsymbol{0}$ and ${n}_L= 1$, \ie zero anchor momentum and fixed Lagrangian value. 
For the force \ac{gp}, we enforce ${\boldsymbol{n}}_F=\boldsymbol{0}$ at $\bar{\boldsymbol{x}}_F = \bar{\boldsymbol{s}}_F = (\mathbf{0}, \mathbf{0}, \mathbf{0})$, which corresponds to $\boldsymbol{F}(\boldsymbol{0}, \boldsymbol{0}, \boldsymbol{0})=\boldsymbol{0}$. 

\paragraph{Test data.}
We evaluate the long-term prediction performance in two scenarios with $5~\mathrm{s}$ length, differing in their input sequences. 
Specifically, 
\begin{itemize}
	\item task 2.1 (input \& dissipation) features a non-zero input $u$, and
	\item task 2.2 (dissipation only) features a zero input.
\end{itemize}
The first task tests the correct interaction between the dynamics, dissipation, and input torques. 
The second task is suited to verify if the system energy is dissipated over time, which is what a physically consistent model would provide. 
For each scenario, predictions are initialized from two successive measured positions and compared against the measured ground truth.

\subsubsection{Real-World Soft Robot} \label{app:rw_soft_robot}

\begin{figure}[tb]
	\centering
	{\fontsize{8pt}{8pt}\selectfont
		\resizebox{0.3\textwidth}{!}{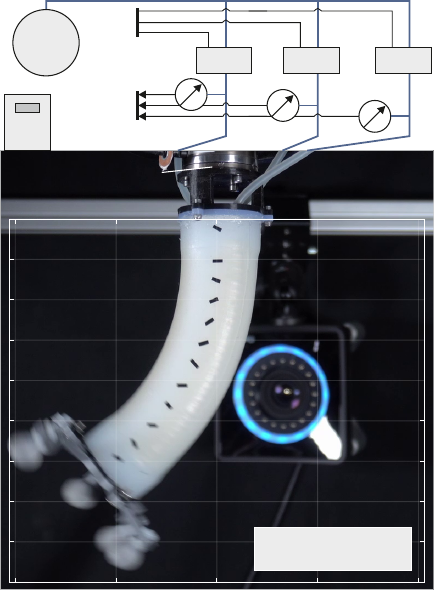}} \hfill
	\includegraphics[width=0.68\textwidth]{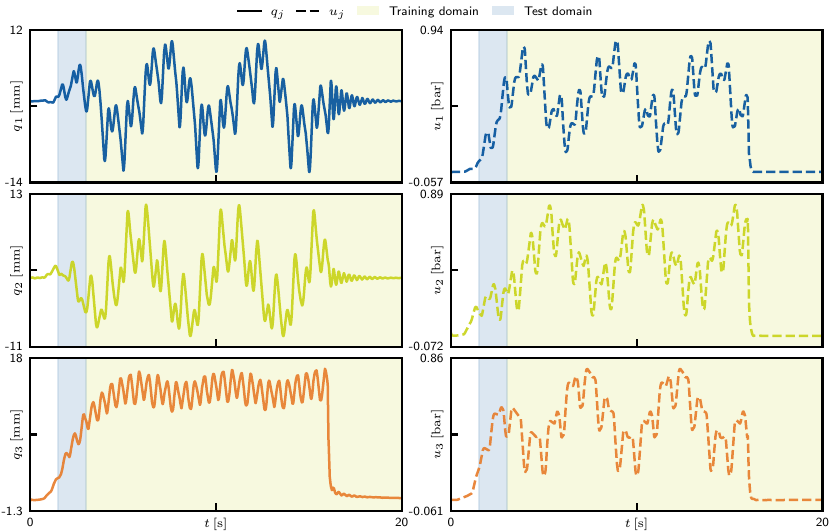}
	\caption{Setup of the pneumatically controlled soft robotic system (left) and employed data set (right). Soft robot image adopted from \cite{Ewering.2026,Mehl.2024}.}
	\label{fig:app_soft_robot_system}
\end{figure}

\paragraph{System setup.}
We consider a controlled soft-robot presented in \cite{Mehl.2022} and \cite{Mehl.2024}. 
The continuous-material soft robot is actuated by air pressure inputs and exhibits nonlinear, hysteretic, and dissipative dynamics. 
The driving input is the chamber pressure $\boldsymbol{u} \in\mathbb{R}^{3}$ and the ``position measurements'' are so-called shape parameters
\begin{equation}\label{eq:app_shape_parameters}
	\boldsymbol{q}^{\top}=\begin{bmatrix}\Delta x & \Delta y & \delta \ell\end{bmatrix}\in\mathbb{R}^{3} \, ,
\end{equation}
obtained from a motion-tracking pipeline. 
Specifically, unlike rigid-body dynamics, the true configuration of a soft continuum robot is effectively infinite-dimensional. 
Still, employing a constant-curvature parametrization as in \cite{DellaSantina.2020}, the robot pose can be represented by the shape parameters \eqref{eq:app_shape_parameters} 
which provide a low-dimensional, physically meaningful state description. 
This choice is compatible with energy-based modeling and has been shown to be useful for physics-informed learning of soft-robot dynamics \cite{Liu.2024}. 
The employed system setup and dataset are displayed in Figure\,\ref{fig:app_soft_robot_system}.

\begin{figure}[tb]
	\centering
	\includegraphics[width=1\textwidth]{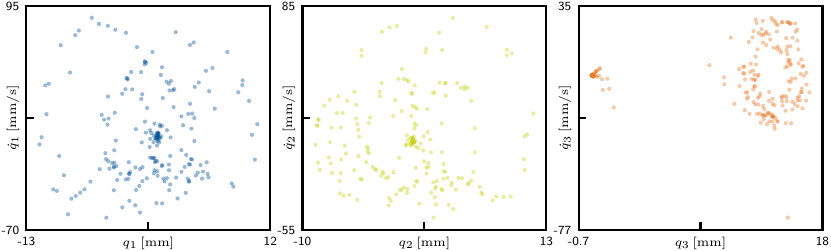}
	\caption{Scatter plot of the randomly drawn training data triplets of the soft robotic system. Each point in the plot represents a triplet.}
	\label{fig:app_soft_robot_system_training_data}
\end{figure}

\paragraph{Training data.}
To generate the training data set, we sub-sample the recorded trajectory by a factor of $20$, which results in a training step size of $\Delta t_{\mathrm{train}} = 20~\mathrm{ms}$. 
We then randomly draw $N=200$ data triplets 
\begin{equation}
	\left\{\boldsymbol{q}_{n-1}^{(i)},\boldsymbol{q}_{n}^{(i)},\boldsymbol{q}_{n+1}^{(i)}\right\}_{i=1}^{N} \, ,
\end{equation}
together with the corresponding pressure inputs. 
A single training data segment of $100$ steps is held out for tuning the scalar slack parameter $\sigma$ in $\boldsymbol{\Sigma} := \sigma \, \mathbf{I}$ (see  \eqref{eq:theta_matrix_expanded}) by a line search. 
The resulting training data distribution is visualized in Figure\,\ref{fig:app_soft_robot_system_training_data}. 

\paragraph{Normalization and kernel design.}
We train both the continuous and discrete \acp{lgp} using a \textit{physics-inspired} kernel design.  
The physics-inspired Lagrangian kernel follows the decomposition in Appendix~\ref{sec:phys_kernel_structure}, that is
\begin{equation}
	k_L = k_M(\boldsymbol{q},\boldsymbol{q}')\,(\dot{\boldsymbol{q}}^{\top}\dot{\boldsymbol{q}}')^2
	+ k_G(\boldsymbol{q},\boldsymbol{q}')
	+ k_S(\boldsymbol{q},\boldsymbol{q}')\,(\boldsymbol{q}^{\top}\boldsymbol{q}')^2 \, ,
\end{equation}
with anisotropic squared exponential sub-kernels $k_M,k_G,k_S$.  
The force kernel is
\begin{equation}
	\bar{k}_F = \boldsymbol{u}^{\top}\boldsymbol{u}'
	+ k_D \left( \begin{bmatrix}
		\boldsymbol{u} \\ \boldsymbol{q}
	\end{bmatrix} , \begin{bmatrix}
		\boldsymbol{u}' \\ \boldsymbol{q}'
	\end{bmatrix} \right) \,(\dot{\boldsymbol{q}}^{\top}\dot{\boldsymbol{q}}') \, ,
\end{equation}
where $k_D$ is a squared exponential kernel. 
Notably, we know from engineering principles that the robot's damping characteristics depend on the input air pressure, as the robot is ``inflated''. 
This knowledge is reflected in the \ac{gp} kernel $k_D$ by taking the positions $\boldsymbol{q}$ \textit{and} the pressure inputs $\boldsymbol{u}$ as features. 
In the discrete \ac{lgp}, velocity factors are approximated as finite differences, $\dot{\boldsymbol{q}}\approx(\boldsymbol{q}_{n+1}-\boldsymbol{q}_n)/\Delta t_{\mathrm{train}}$, which aligns with the employed midpoint rule in Section\,\ref{sec:ct_lgp}. 

The normalization conditions are imposed as described in Section~\ref{sec:dt_lgp}. 
For the Lagrangian, we use the anchor point $\bar{\boldsymbol{z}}_L = \bar{\boldsymbol{r}}_L = (\mathbf{0}, \mathbf{0})$  with targets $\boldsymbol{n}_M= \boldsymbol{0}$ and ${n}_L= 1$, \ie zero anchor momentum and fixed Lagrangian value. 
For the force \ac{gp}, we enforce ${\boldsymbol{n}}_F=\boldsymbol{0}$ at $\bar{\boldsymbol{x}}_F = \bar{\boldsymbol{s}}_F = (\mathbf{0}, \mathbf{0}, \mathbf{0})$, which corresponds to $\boldsymbol{F}(\boldsymbol{0}, \boldsymbol{0}, \boldsymbol{0})=\boldsymbol{0}$.

\paragraph{Test data.}
Testing is performed on a held-out validation trajectory segment (see Figure\,\ref{fig:app_soft_robot_system}).
The prediction rollouts are initialized from two successive measured positions and compared against the ground truth over the horizon.

\subsection{Evaluation Metrics} \label{app:metrics}

\textbf{\acl{rmse}} (\acs{rmse}) measures the average magnitude of the errors between the predicted trajectories and the ground truth. 
A lower \acs{rmse} indicates a more accurate prediction. The \acs{rmse} is defined as
\begin{equation}
\mathrm{RMSE} = 
    \sqrt{
      \frac{1}{n_q \,N_{\mathrm{pred}}}
      \sum_{i=1}^{n_q}
      \sum_{j=1}^{N_{\mathrm{pred}}}
        \bigl(q_{i,j}^{\mathrm{true}} - {q}_{i,j}\bigr)^2
    } \, ,
\end{equation}

where $n_q$ is the total number of generalized coordinates, $N_\mathrm{pred}$ is the number of prediction steps, $q_{i,j}^{\mathrm{true}}$ is the ground-truth and ${q}_{i,j}$ is the prediction for the $i$-th dimension and $j$-th prediction step.

\section{Additional Experimental Results}\label{app:additional_results}

We present here additional experimental results. First, in Appendix\,\ref{app:case_studies_pendulum}, we give the full results of the quantitative study with the multi-body pendulums. 
Second, in Appendix\,\ref{app:pendulum_long_term_pred}, we examine energy conservation in a simulation example. 
Third, we provide a visualization of Task\,2.1 in the real-world double pendulum (see Appendix\,\ref{app:case_studies_rw_pendulum}). 
Last, we give an additional case study in a non-harmonic oscillator system with two equilibria in Appendix\,\ref{app:case_studies_sim_non_harmonic_oscillator}.

\subsection{Extended Quantitative Study with Multi-Link Pendulum}\label{app:case_studies_pendulum}

In this appendix, we give the full quantitative study with multi-body pendulum simulations. 
Specifically, we consider a standard \ac{gp} and discrete/continuous \acp{lgp} with \textit{generic} and \textit{physics-inspired} kernels in controlled single, double, and triple pendulums, \ie $n_q \in \{1,2,3\}$. 
We train all methods at the training and prediction step size $\Delta t_{\mathrm{train}}=\Delta t_{\mathrm{pred}}=5 \cdot 10^{-2}$ for different training data budgets $N \in \{25, 50, 100, 200, 300\}$. 
For evaluation, we use the \ac{rmse} between a ground-truth $20$-step simulation and the corresponding learning-based predictions.

\begin{figure}[tb]
	\centering
	\includegraphics[width=1\textwidth]{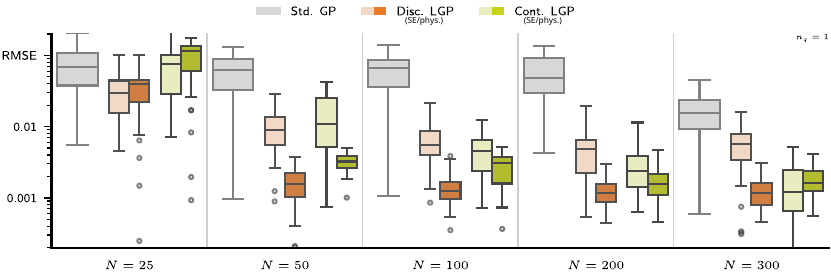}
	\includegraphics[width=1\textwidth]{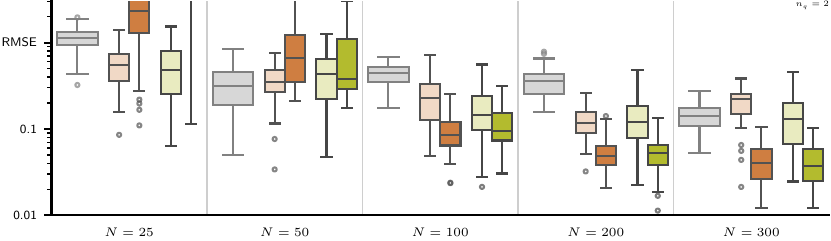}
	\includegraphics[width=1\textwidth]{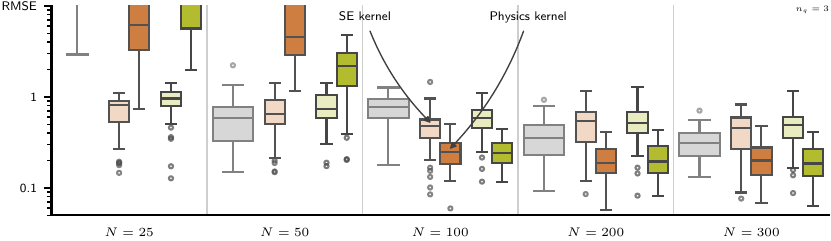}
	\caption{Error against training data budget in $20$-step simulations ($n_q \in \{1,2,3\}$, $50$ random simulations).}
	\label{fig:sim_pendulum_n_train}
\end{figure}

The results of the quantitative study are visualized in Figure\,\ref{fig:sim_pendulum_n_train}. 
As expected, adding prior knowledge about the data-generating dynamics improves the predictive performance. 
First, we see that the proposed \acp{lgp} outperform the standard \ac{gp} predictions across dimensions. 
Second, the discrete and continuous \acp{lgp} with physics-inspired kernel provide more accurate predictions than \acp{lgp} with generic kernels. 
This methodological characteristic is promising, as the employed knowledge that an embodied system obeys the Lagrange-d'Alembert principle is very general, meaning that only \emph{few system-specific expert knowledge} is required. 
Therefore, the \acp{lgp} provide a vastly improved predictive performance essentially \emph{for free} for the considered system class.

Given extremely sparse data, the \acp{lgp} achieve performance comparable to the standard \ac{gp} baseline. 
The improvement relative to standard \acp{gp} is particularly evident for system dimensions $n_q > 1$ and from approximately $N=100$ data points. 
In this regime, we observe that \acp{lgp} trained with $N=100$ data points perform \emph{on par} or better than a standard \ac{gp} trained on three times more data points (see Figure\,\ref{fig:sim_pendulum_n_train} for $n_q=2$). 
We hypothesize that this behavior is due to the indirect learning objective of satisfying \eqref{eq:DEL}, and potential identifiability issues in kernel architectures with multiple sub-components (see Appendix\,\ref{sec:phys_kernel_structure}).

\newpage
\subsection{Structure Preservation in Long-term Predictions}\label{app:pendulum_long_term_pred}

\begin{figure}[tb]
	\centering
	\includegraphics[width=1\textwidth]{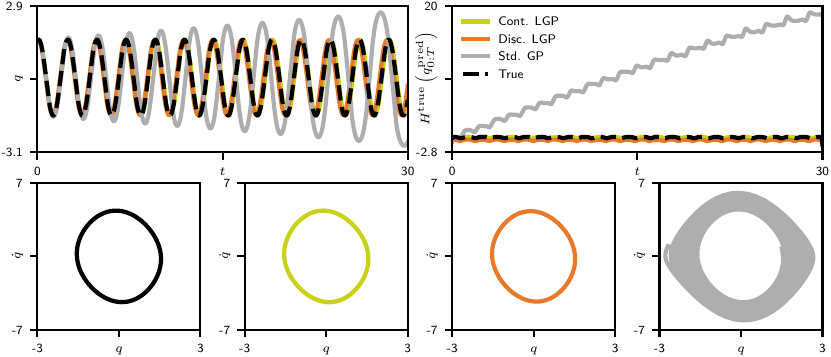}
	\caption{Long-term forward predictions of position $q$ (top left) and system energy $H$ (top right) in a conservative single pendulum, \ie without dissipation or inputs. In the bottom plots, the associated trajectories are shown in phase space. The true system does not present an energy drift under a variational integration scheme. The \acp{lgp} match this property by preserving the geometric structure of the underlying Lagrange-d'Alembert principle by construction. The predictions of a standard \ac{gp} show a drift in positions and energy.}
	\label{fig:sim_pendulum_energy_drift}
\end{figure}

The proposed \acp{lgp} preserve the geometric structure of the Lagrange-d'Alembert principle by construction in the absence of external forces. 
We illustrate this property in Figure\,\ref{fig:pendulum_2_structure_preservation} and give further details on the implications for long-term predictions in this appendix. 
Specifically, we simulate a conservative single pendulum, \ie with zero inputs and damping coefficient $b=0$, forward for $30$ seconds at time step width $\Delta t_{\mathrm{train}}=\Delta t_{\mathrm{pred}}= 5 \cdot 10^{-2}$ and evaluate the predictive performance of different learning-based dynamics models. 
In Figure\,\ref{fig:sim_pendulum_energy_drift}, we compare the long-term forward predictions of \acp{lgp} and standard \ac{gp} in terms of their position coordinate predictions $q$ and the associated system energy $H$. 
To construct the current system energy $H$, we evaluate the true Hamiltonian at the predicted trajectory samples. 
In the example, the \acp{lgp} use a physics-inspired kernel. 

The evaluation in Figure\,\ref{fig:sim_pendulum_energy_drift} shows that the \acp{lgp} yield accurate and stable long-term predictions with a system energy that oscillates around a constant level. 
As discussed in Section\,\ref{sec:results_sim_pendulum}, the oscillation is characteristic of the employed variational integration procedure and occurs in the ground truth solution as well. 
In contrast, the standard \ac{gp} position predictions diverge, leading to a physically inconsistent accumulation of system energy over time.

\subsection{Controlled Real-World Double Pendulum with Inputs and Dissipation}\label{app:case_studies_rw_pendulum}

\begin{figure}[tb]
	\centering
	\includegraphics[width=1\textwidth]{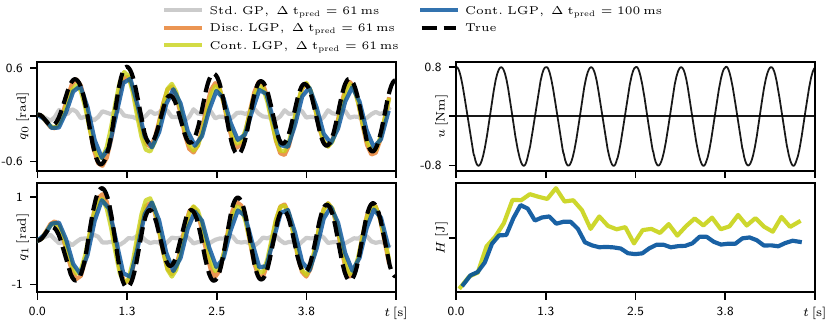}
	\caption{Prediction task 2.1 \textit{with inputs and dissipation} in a controlled real-world double pendulum. The \acsp{lgp} yield accurate forward simulations despite learning only from noisy real-world position data ($\Delta t_{\mathrm{train}}=61\,\mathrm{ms}$, $n_q =2$, $N=300$ data points).}
	\label{fig:real_world_pendulum_app}
\end{figure}

In the real-world double pendulum case study, we investigate two prediction tasks, one with control inputs (Task\,2.1) and one without (Task\,2.2). 
We summarize the prediction results and visualize Task\,2.2 in Figure\,\ref{fig:real_world_pendulum} of the main paper. 
Here, in Figure\,\ref{fig:real_world_pendulum_app}, we give the results of Task\,2.1.

\newpage
\subsection{Non-Harmonic Oscillator with Multiple Equilibria}\label{app:case_studies_sim_non_harmonic_oscillator}

\begin{wrapfigure}{r}{0.4\textwidth} 
    \vspace{-5mm}
    \centering
    {\fontsize{6pt}{6pt}\selectfont
        \resizebox{0.4\textwidth}{!}{
            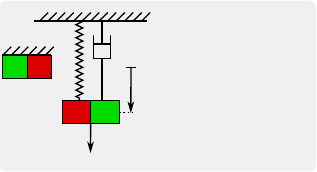
        }
    }
    \vspace{-6mm}
    \caption{Non-harmonic oscillator and its Hamiltonian.}
    \label{fig:app_nonharmonic_oscillator}
    \vspace{-5mm}
\end{wrapfigure} 

In this appendix, we give an additional simulation study of the non-harmonic oscillator depicted in Figure\,\ref{fig:app_nonharmonic_oscillator}. 
This example is interesting because it features two distinct equilibria that do not lie at the origin. 
Specifically, we consider the motion of the oscillator's positional coordinate $q$ that is governed by the Lagrangian 
\begin{equation}
    L(q, \dot{q}) = \frac{1}{2} m \dot{q}^2 - \left( \frac{1}{4} k q^2 + c \cos(q) \right) \, .
\end{equation}
The first term in the Lagrangian function represents the kinetic energy, while the bracketed terms define a non-harmonic potential, originating from the static magnet that pushes a spring-loaded magnet either into an upper or lower equilibrium. 
Additionally, the system is subjected to a generalized non-conservative external force given by
\begin{equation}
    F(u, q, \dot{q}) = u - b \dot{q} \, ,
\end{equation}
which incorporates an external control input $u$ and linear viscous damping dependent on the velocity. 
The simulation features four scalar parameters: the mass $m:=1$, the ``stiffness'' coefficient $k:=1$, the magnetic potential energy $c:=2$, and the linear damping coefficient $b:=0.1$. 

\paragraph{Training.} We compare two continuous \acp{lgp} with \emph{generic} squared exponential kernel and \emph{physics-inspired} kernel. 
In both cases, we use the same kernels as in the multi-link pendulum case study in Section\,\ref{sec:results_sim_pendulum} and Appendix\,\ref{app:multi_link_pendulum}. 
For training, we sample $N=500$ position data triplets at a time step width of $\Delta t_{\mathrm{train}} = \Delta t_{\mathrm{pred}}= 5 \cdot 10^{-2} $. 

\begin{figure}[tb]
	\centering
	\includegraphics[width=1\textwidth]{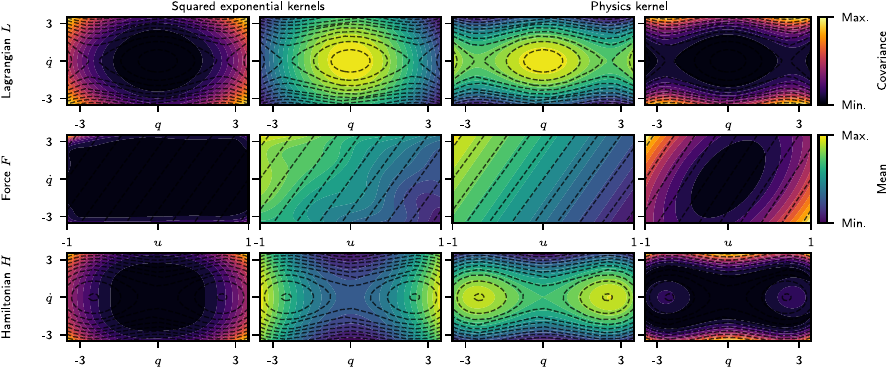}
	\caption{Learned quantities in the non-harmonic oscillator example ($N=500$ data points, $n_q=1$, dashed: true $L$, $F$, and $H$).}
	\label{fig:sim_nonharmonic_oscillator}
\end{figure}

\paragraph{Learned dynamics model.} 
{First}, we are interested in learning the external force $F$, the Lagrangian $L$, and the Hamiltonian $H$ of the oscillator. 
Figure\,\ref{fig:app_nonharmonic_oscillator} summarizes the learned quantities. 
As expected, the \ac{lgp} with physics-inspired kernel matches the true quantities closely, while the variant with a generic kernel is less precise. 
Still, both methods capture the overall system characteristic of having two attractive regions along the $q$-axis.

\begin{figure}[tb]
	\centering
	\includegraphics[width=0.63\textwidth]{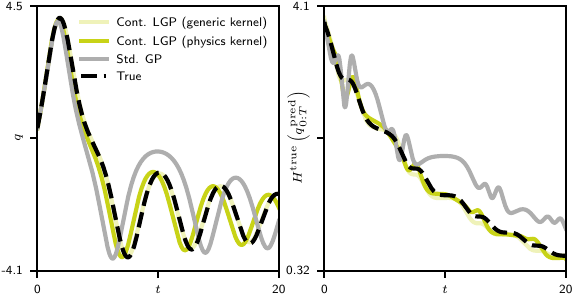}\hfill
    \includegraphics[width=0.36\textwidth]{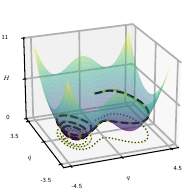}
	\caption{Predicted position and energy trajectories in the non-harmonic oscillator example over time (left) and along the true Hamiltonian function (right).}
	\label{fig:sim_nonharmonic_oscillator_trajectories}
\end{figure}

\paragraph{Forward predictions.} 
Second, we are interested in performing multi-step forward predictions. 
As evident in Figure\,\ref{fig:sim_nonharmonic_oscillator_trajectories}, both \acp{lgp} provide accurate long-term predictions, correctly converging to one of the equilibria, and outperform a baseline standard \ac{gp}. 
In this case study, the continuous \ac{lgp} with a generic squared exponential kernel provides more accurate predictions than the physics-inspired version. 
We hypothesize that this behavior is due to more free kernel hyperparameters to optimize in the physics-kernel case or due to less approximative flexibility.

\newpage
\section{Computational Resources and Software}\label{app:miscellanea}

\paragraph{Computational resources.} 
All experiments presented in this work are performed on a standard office computer, equipped with an Intel Core i5-1235U (1.30 GHz) and 8 GB RAM. 
Representative training and prediction scenarios with the \acp{lgp} take a few seconds to minutes, depending on the number of training data points and the hyperparameter optimization settings. 
The memory requirements are comparable to those of standard \acp{gp}. 

\paragraph{Software.} 
The code base is built upon \texttt{JAX} (\url{https://docs.jax.dev/en/latest/}, License: Apache-2.0). 

\paragraph{Data.} 
The employed figure and data of the real-world double pendulum are provided by the \texttt{DFKI RIC Underactuated Robotics Lab} via GitHub (\url{https://github.com/dfki-ric-underactuated-lab/double_pendulum.git}, License: BSD 3-Clause). 
The authors have the exclusive right to use the dataset of the real-world pneumatic soft robot. 

\section{Use of Large Language Models}
We utilized \acp{llm} throughout multiple phases of this study. During the early stages, they aided in brainstorming methodologies and conducting literature reviews. 
As the project progressed, \acp{llm} functioned as coding assistants, helping to write and debug algorithms. 
Finally, we employed these tools to refine the manuscript, polishing its grammar, clarity, and overall flow.


%

\end{document}